\RequirePackage{fix-cm}
\documentclass[twocolumn]{svjour3}          
\smartqed  
\usepackage{graphicx}
\usepackage{hyperref}       
\usepackage{url}            
\usepackage{booktabs}       
\usepackage{amsfonts}       
\usepackage{nicefrac}       
\usepackage{microtype}      
\usepackage{flushend}
\usepackage{multirow}
\usepackage{multicol}
\usepackage{amssymb}
\usepackage{amsmath}
\usepackage{subcaption}
\usepackage{colortbl}
\usepackage{changepage}
\usepackage{soul, color}

\usepackage[dvipsnames]{xcolor}

\journalname{International Journal of Computer Vision}
\begin{document}

\title{HiEve: A Large-Scale Benchmark for Human-centric Video Analysis in Complex Events}


\author{Weiyao Lin \and Huabin Liu \and Shizhan Liu \and Yuxi Li \and  Hongkai Xiong \and Guojun Qi \and Nicu Sebe
}


\institute{
           Weiyao Lin \and Huabin Liu \and Shizhan Liu \and Yuxi Li \and Hongkai Xiong \at
			Department of Electronic Engineering, Shanghai Jiao Tong University, Shanghai China \\
			\email{\{wylin,huabinliu,shanluzuode,lyxok1,xionghongkai\} @sjtu.edu.cn}	
            \and
            Guojun Qi \at
            Machine Perception and Learning Lab, USA \\
            \email{guojunq@gmail.com}
            \and
            Nicu Sebe \at
            University of Trento, Trento, Italy \\
            \email{niculae.sebe@unitn.it}
            \and
}

\date{Received: date / Accepted: date}

\maketitle

\begin{abstract}
  Along with the development of modern smart cities, human-centric video analysis has been encountering the challenge of analyzing diverse and complex events in real scenes. A complex event relates to dense crowds, anomalous individuals, or collective behaviors. However,  limited by the scale and coverage of existing video datasets, few  human analysis approaches have reported their performances on such complex events. To this end, we present a new large-scale dataset with comprehensive annotations, named Human-in-Events or HiEve (Human-centric video analysis in complex Events), for the understanding of human motions, poses, and actions in a variety of realistic events, especially in crowd \& complex events. It contains a record number of poses (\textgreater 1M), the largest number of action instances (\textgreater56k) under complex events, as well as one of the largest numbers of trajectories lasting for longer time (with an average trajectory length of \textgreater480 frames). Based on its diverse annotation, we present two simple baselines for action recognition and pose estimation, respectively. They leverage cross-label information during training to enhance the feature learning in corresponding visual tasks. Experiments show that they could boost the performance of existing action recognition and pose estimation pipelines. More importantly, they prove the widely ranged annotations in HiEve can improve various video tasks. Furthermore, we conduct extensive experiments to benchmark recent video analysis approaches together with our baseline methods, demonstrating HiEve is a challenging dataset for human-centric video analysis. We expect that the dataset will advance the development of cutting-edge techniques in human-centric analysis and the understanding of complex events. The dataset is available at \href{http://humaninevents.org}{\textcolor{magenta}{http://humaninevents.org}}. 
\end{abstract}
 

\section{Introduction}
The development of smart cities highly relies on the advancement of fast and accurate visual understanding of multimedia~\cite{xumai,meitao1,chen2022audio}. To achieve this goal, many human-centered and event-driven visual understanding problems have been raised, such as human pose estimation~\cite{alpha2017}, pedestrian tracking~\cite{mot2020,lujiwen}, and action recognition \cite{veeriah2015differential,shu2019hierarchical}.

Recently, several public datasets (e.g., MSCOCO~\cite{coco}, PoseTrack~\cite{posetrack2018}, UCF-Crime~\cite{ucf-crime}) have been proposed to benchmark the aforementioned tasks. 
However, they have some limitations when applied to real scenarios with complex events such as dining, earthquake escape, subway getting-off and collisions.
\textit{\textbf{First}}, most benchmarks focus on normal or relatively simple scenes. These scenes either have few occlusions or contain many easily-predictable motions and poses. \textit{\textbf{Second}}, the coverage and scale of existing benchmarks are still limited. For example, although the UCF-Crime dataset~\cite{ucf-crime} contains challenging scenes, it only has coarse video-level action labels which may not be enough for fine-grained action recognition of human instance. Similarly, although the numbers of pose labels in MSCOCO~\cite{coco} and PoseTrack~\cite{posetrack2018} are sufficiently large for simple scenes with limited occlusions, these datasets lack realistic scenes containing crowded scenes and complex events.

To this end, we present a new large-scale human-centric dataset, named Human-in-Events (HiEve), for understanding a hierarchy of human-centric information (motions, poses, and actions) in a variety of realistic complex events, especially in crowded and complex events. Among all datasets for realistic crowd scenarios, HiEve has substantially larger scales and complexity and contains a record number of poses (\textgreater1M), action labels (\textgreater56k) and long trajectories (with average trajectory length \textgreater480 frames). Compared with existing datasets, HiEve contains more comprehensive and larger-scale annotations in both generic and complex scenes, making it more  
adequate to develop new human-centric analysis techniques and evaluate them in realistic scenes. \autoref{table:comprision} provides a quantitative comparison of the HiEve dataset with related datasets in light of their nature and scale. 

One main feature of our HiEve dataset is the hierarchical and diverse information of human annotations under unified crowd scenes, which encourages to accomplish multiple human-centric visual tasks by integrating cross-annotation information. To make a tentative validation of this property, we explore combining the pose and action label in HiEve to present two simple baselines (1) a pose-aware action recognition algorithm and (2) an action-guided pose estimation algorithm. Specifically, the former promotes video action learning by encouraging the video feature to predict pose-aware motion patterns, while the latter refines the pose representation with action category prior knowledge. Experiments demonstrate that they can boost the performance of existing state-of-the-art pipelines on our HiEve dataset. We hope this exploration will foster further research in video understanding with diverse annotations of HiEve. 

Additionally, we build an online evaluation server available to the whole community in order to enable timely and scalable evaluation on the held-out test videos. 
We also evaluate existing state-of-the-art solutions on HiEve to benchmark their performance and analyze the corresponding oracle models, demonstrating that HiEve is challenging and of great value for advancing human-centric video analysis.
In summary, we make the following main contributions:
\begin{itemize}
	\item We collect a new large-scale video dataset HiEve under various realistic complex events (e.g., dining, earth-quake escape, collision) for human-centric video analysis.
	\item Our HiEve provides a wide range of human annotations (track, pose, action) to enable analysis on various visual tasks, such as multi-object tracking, pose estimation, and action recognition.
	\item By virtue of the diverse annotation in HiEve, we propose two enhanced baselines for action recognition and pose estimation, respectively. Experiments on them demonstrate the correlation between different types of human annotations could further boost the state-of-the-art methods on HiEve.
\end{itemize}

\begin{table*}[ht]
\centering
\resizebox{\linewidth}{!}{
\begin{tabular}{ccccccccc}
	\hline
	\textbf{Dataset}  & \textbf{\# pose}  & \textbf{\# box} & \textbf{\# traj.(avg)} & \textbf{\# action(class)} & \textbf{\# total length (avg)} & \textbf{pose track} & \textbf{surveillance} & \textbf{complex events} \\ \hline
	MSCOCO~\cite{coco}          & 105,698            & 105,698         & NA              & NA      & NA          & $\times$                   & $\times$                   & $\times$            \\
	MPII~\cite{mpii2014}            & 14,993             & 14,993          & NA              & 25,000       & NA        & $\times$                   & $\times$                     & $\times$                       \\
	CrowdPose~\cite{crowdpose2019}       & $\sim$80,000       & $\sim$80,000    & NA              & NA       & NA         & $\times$                   & $\times$                     & $\times$                       \\
	PoseTrack~\cite{posetrack2018}      & $\sim$267,000      & $\sim$26,000    & 5,245(49)       & NA       & 2,750s(2s)        & $\surd$                   & $\times$                     & $\times$                       \\
	MOT16\cite{mot16}           & NA                 & 292,733         & 1,276(229)      & NA       & 463s(33s)       & $\times$                   & $\surd$                     & $\times$                       \\
	MOT17          & NA                 & 901,119         & 3,993(226)      & NA       & 1,389s(66s)         & $\times$                   & $\surd$                     & $\times$                       \\
	MOT20~\cite{mot2020}           & NA                 & 1,652,040       & 3457(478)      & NA       & 535s(67s)         & $\times$                   & $\surd$                     & $\times$                       \\
	Avenue~\cite{avenue2013}          & NA                 & NA              & NA              & 37(37)       & 1,225s(33s)      & $\times$                   & $\surd$                     & $\times$                       \\
	UCF-Crime~\cite{ucf-crime}       & NA                 & NA              & NA              & 1,900(13)      & 128h(4s)       & $\times$                   & $\surd$                     & $\surd$                       \\
	UCF101-24~\cite{ucf101_2012}       & NA                 & NA              & NA              & 44,716(\textbf{24})      & $\sim$4h(7s)       & $\times$                   & $\times$                     & $\times$                       \\
	JHMDB-21~\cite{hmdb2011}       & NA                 & NA              & NA              & 31,838(21)      & $\sim$5h(9s)       & $\times$                   & $\times$            &         $\times$                      \\				
	\textbf{HiEve (Ours)}   & \textbf{1,099,357} & 1,302,481       & 2,687(485)      & \textbf{56,643}(14)  & 1,839s(57s) & \textbf{$\surd$}          & \textbf{$\surd$}            & \textbf{$\surd$}              \\ \hline
\end{tabular}}
\caption{Comparison between HiEve and existing datasets. ``NA'' indicates not available. ``$\sim$'' denotes approximated value. For ``traj.(avg)'', the ``traj.'' means trajectory and ``avg'' indicates average trajectory length. For ``action(class)'', ``action'' means action instance and ``class'' indicates the number of action category. For ``total length (avg)'', ``total length'' denotes the total length of all videos while the ``avg'' means the average video length.}
\label{table:comprision}
\end{table*}

\section{Related benchmarks and Comparison}
\subsection{Multi-object Tracking Datasets}
Different from single-object tracking, multi-object tracking (MOT) does not solely depend on sophisticated appearance models to track objects in frames.  In recent years, there is a corpus of datasets that provide multi-object bounding-box and track annotations in video sequences, which have fostered the development of this field. PETS~\cite{pets2009} is an early proposed multi-sensor video dataset, it includes annotation of crowd person count and tracking of an individual within a crowd. Its sequences are all shot in the same scene, which leads to relatively simple samples. KITTI~\cite{kitti2012} tracking dataset features videos from a vehicle-mounted camera and focuses on street scenarios, it owns 2D \& 3D bounding-boxes and tracklets annotations. Meanwhile, is has a limited variety of video angles. The MOT-Challenge dataset~\cite{mot16} is the most widely-used benchmark for MOT tasks, primarily focusing on evaluating tracking performance in crowded environments.     While the MOT-series (MOT-17, 19, and 20) datasets have fostered the development of various tracking algorithms, they exhibit certain shortcomings for current real-world applications.     A key limitation of the MOT-Challenge dataset is its relatively narrow scope, as it predominantly features scenes with pedestrians in urban settings.  This lack of diversity in scene types and events may hinder the generalization of tracking algorithms to more complex and varied scenarios.  Compared to the latest MOT-20~\cite{mot2020} dataset, our HiEve dataset collects videos from various real-world scenes (12 scenes in total) and includes more complex events, such as fights, earthquakes, and robberies, presenting more significant challenges for real-world MOT tasks.     Furthermore, as shown in \autoref{table:comprision}, HiEve has longer video and track lengths than MOT20.  Most importantly, HiEve offers a broad range of annotations, encompassing dense human poses, object tracking, and actions, making it a more comprehensive dataset for human-centric understanding tasks.

\subsection{Pose Estimation and Tracking Datasets}
Human pose estimation in images has made great progress over the last few years. For single-person pose estimation, LSP~\cite{lsp2010}, FLIC~\cite{filc2013} are the two most representative benchmarks, the former focuses on sports scenes while the latter is collected from popular Hollywood movie sequences. Compared with LSP, FLIC only labels 10 upper body joints and owns a smaller data scale.

WAF~\cite{waf2010} is the first to establish a benchmark for multi-person pose estimation with simplified keypoint and body definition. Then, MPII~\cite{mpii2014} and MSCOCO~\cite{coco} datasets were proposed to further advance the multi-person pose estimation task by their diversity and difficulty in the human pose. In particular, MSCOCO is regarded as the most widely used large-scale dataset with 105698 pose annotations in hundreds of 
daily activities. To evaluate the performance under crowded scenes, Crowdpose~\cite{crowdpose2019} selects crowded images from MPII, MSCOCO to form a subset for pose estimation under crowded scenes. Therefore, the scale of Crowdpose dataset is limited.
Taking the tracking task into consideration, PoseTrack~\cite{posetrack2018} builds a new video dataset which provides multi-person pose estimation and articulated tracking annotations. Compared with them, our HiEve provides more realistic scenarios for both pose estimation and pose tracking. Meanwhile, HiEve is dominated by crowded scenes, which is more challenging for current pose estimation algorithms.

\subsection{Action Recognition Datasets}
In recent years, action recognition has emerged as a popular research topic in computer vision.    Meanwhile, the availability of large-scale video datasets has greatly facilitated the development of this field.           UCF101~\cite{ucf101_2012} and HMDB-51~\cite{hmdb2011} are two widely used datasets, which consist of various sports videos and daily activities collected from movies and online resources.     The Kinetics~\cite{kinetics2017} dataset, with 400/600/700 action categories and more than 300,000 clips, is currently one of the largest video datasets for action recognition.            Researchers often use this dataset to provide prior action knowledge for downstream video backbones and tasks.     The Epic-Kitchen~\cite{epic-kitchen} and Something-Something~\cite{ssv2} datasets are unique in that they focus on human-object interactions and first-person visions.       Epic-Kitchen collects videos in a daily kitchen setting, while Something-Something focuses on videos that record people performing actions with certain objects.       Both of them pose new and significant challenges for action recognition.     
To recognize the anomaly actions, Avenue~\cite{avenue2013} and UCF-Crime~\cite{ucf-crime} are further proposed.       Aveue collects 37 videos with  abnormal events from the campus, while UCF-Crime annotates 13 anomalies in real-world surveillance videos, such as fighting, accident, and robbery.    However, most of the above datasets are collected from either less realistic drama scenes or uncrowded scenarios. 
	
The benchmarks mentioned above follow the regular video-level action recognition task, where each video is assigned only one action label. However, the action recognition task in our HiEve dataset focuses on a more complicated action detection task, where both the location and category of the action need to be recognized for each object. The previous action detection benchmarks, UCF101-24~\cite{action_detection,action_detection2} and JHMDB-21~\cite{action_detection,action_detection2}, are more similar to our setting. The UCF101-24 dataset is a subset of the UCF101~\cite{ucf101_2012} dataset, focusing specifically on 24 human action classes related to sports and human movements. This subset is annotated with spatiotemporal bounding boxes, making it suitable for the evaluation of both action recognition and action detection tasks. Similar to UCF101-24, the JHMDB-21 is a subset of the large HMDB~\cite{hmdb2011} dataset, which selects 21 classes and annotate them with spatiotemporal bounding boxes. As shown in \autoref{table:comprision}, compared to them, we contains a larger number of action instances and a much longer average video length (57s). Most importantly, the actions in HiEve dataset are performed under complex scenarios or abnormal events, which poses more significant challenges for action detection.

\subsection{HiEve vs. other datasets}
In summary, the related datasets mentioned above have served the community very well, but now they are confronting several limitations: (1) Most of them are focusing on normal or simple scenes (2) Their coverage and scales are limited. (3) They only contain a single aspect of human annotation (pose, track or action). Overall, compared with these datasets, our dataset has the following unique characteristics:
	\begin{itemize}
		\item HiEve dataset covers \textbf{a wide range of human-centric annotations} including track, pose, and action, while the previous datasets only focus on a subset of our tasks.
		\item HiEve dataset focuses on the challenging scenes under \textbf{crowded and complex events} (such as dining, earthquake escape, subway getting-off, and collision), while the previous datasets are mostly related to normal or relatively simple scenes.   
		\item HiEve dataset has substantially \textbf{larger data scales and coverage}, including the currently largest number of poses (\textgreater1M), the largest number of complex-event action labels (\textgreater56k), and one of the largest number of trajectories with long terms (with average trajectory length \textgreater480 frames).
	\end{itemize}
	In a nutshell, our HiEve contains more comprehensive and larger-scale annotations in various complex-event scenes, making it more capable of evaluating the human-centric analyzing techniques in realistic scenes.
	\begin{figure}[t]
		\centering
		\includegraphics[width=0.48\textwidth]{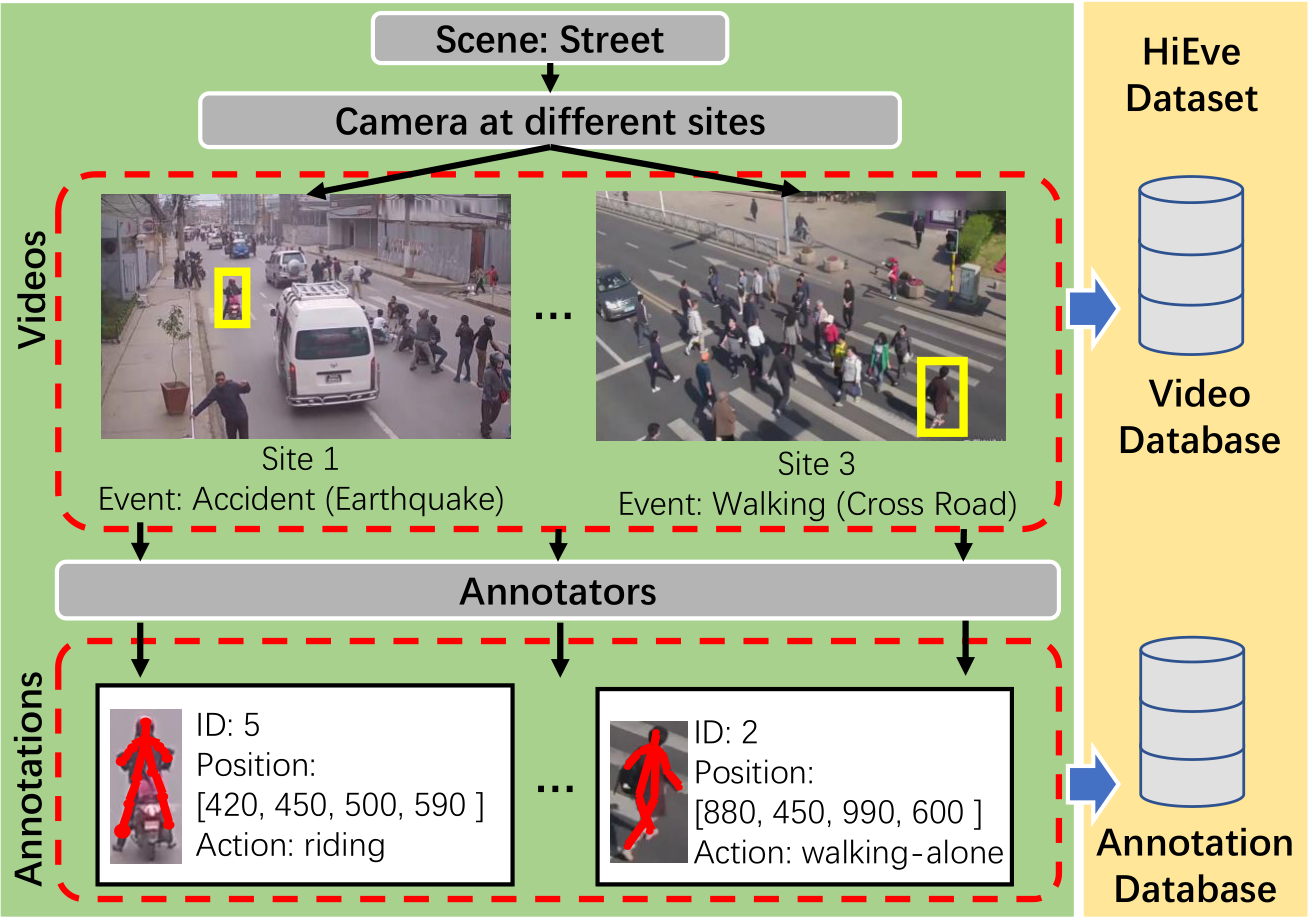}
		\caption{An example of the collection workflow of our HiEve dataset under street scene, where each scene contains videos captured at different sites with different types of events happening.}
		\label{fig:collection}
	\end{figure}

\section{The HiEve dataset}

\begin{figure*}[ht]
	\centering
	\includegraphics[width=\linewidth]{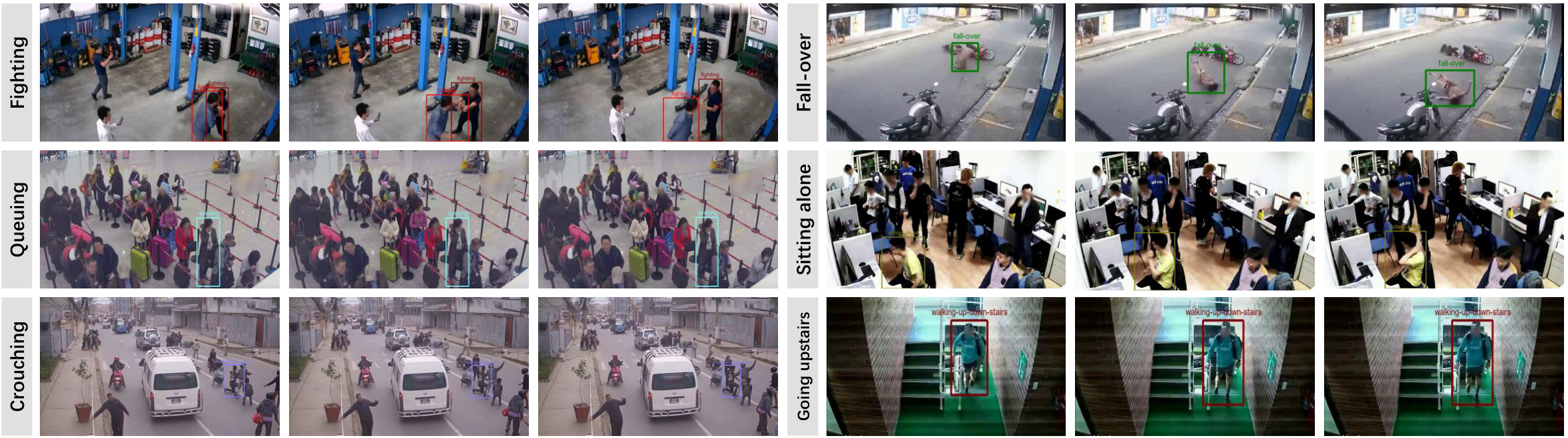}
	\caption{Samples of different actions from our training set and testing set. }
	\label{fig:action_sample}
\end{figure*}

\begin{figure}[t]
	\centering
	\begin{minipage}[t]{0.2\linewidth}
		\centering
		\includegraphics[width=1\textwidth,height=2.82\linewidth]{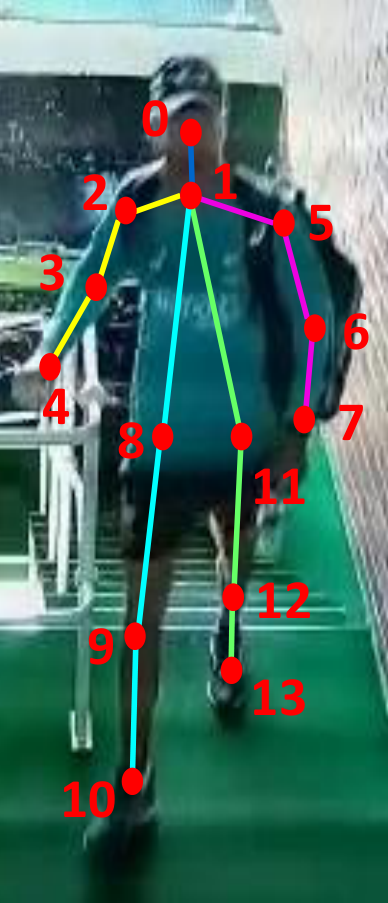}
		\subcaption{}
		\label{fig:keypoints}
	\end{minipage}
	{\vrule width0.6pt}
	\begin{minipage}[t]{0.75\linewidth}
		\centering
		\includegraphics[width=1\textwidth]{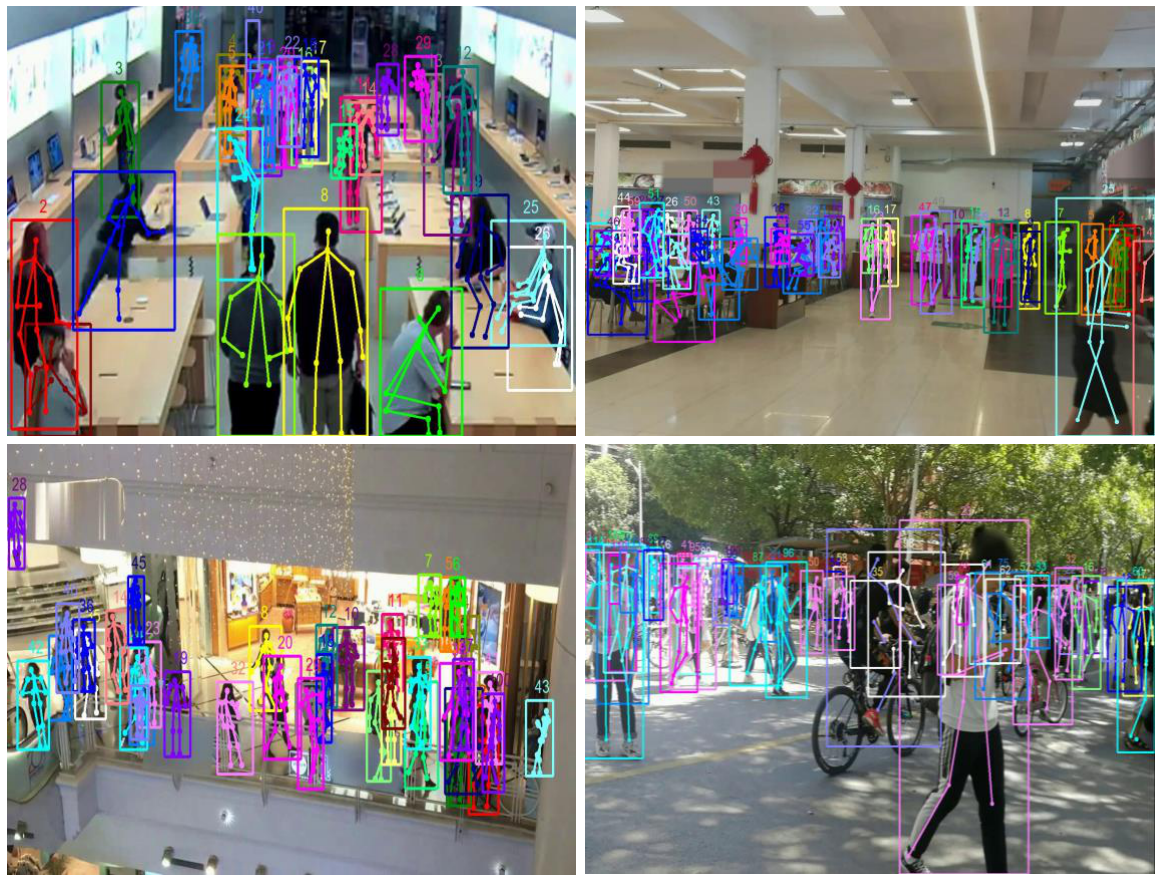}
		\subcaption{}
		\label{fig:pose_sample}
	\end{minipage}  
	\caption{(a)~Keypoints definition (b)~Example pose and bounding-box annotations from our dataset.}
\end{figure}
\subsection{Collection and Annotation}
\begin{figure*}[ht]
	\centering
	\begin{minipage}[t]{0.25\linewidth}
		\vspace{-5cm}
		\includegraphics[width=1.0\textwidth]{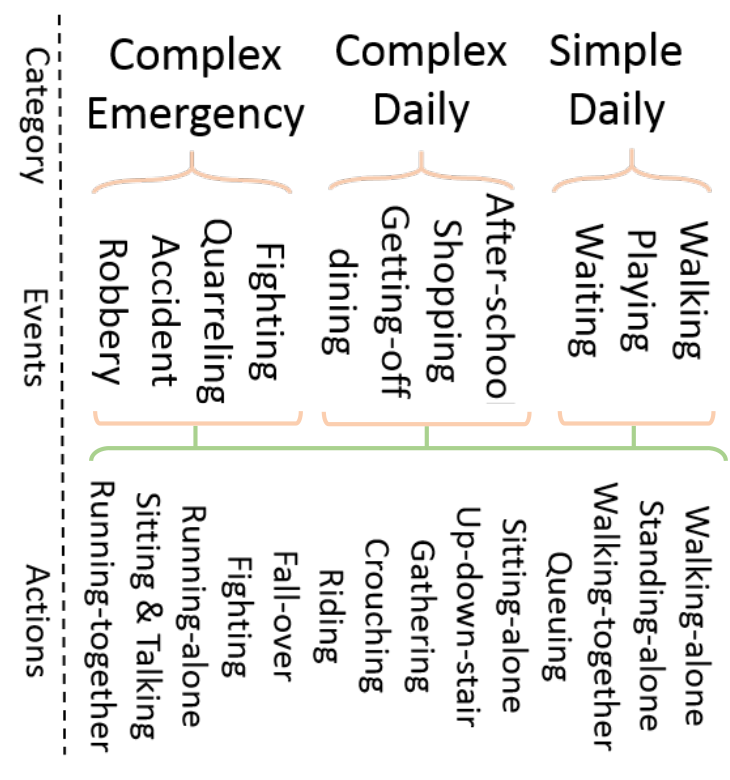}
		\caption{The classification of events. They are divided into three event categories.}
		\label{table:sub-events}
	\end{minipage}
	\hspace{0.5cm}
	\begin{minipage}[t]{0.32\linewidth}
		\centering
		\includegraphics[width=1\textwidth]{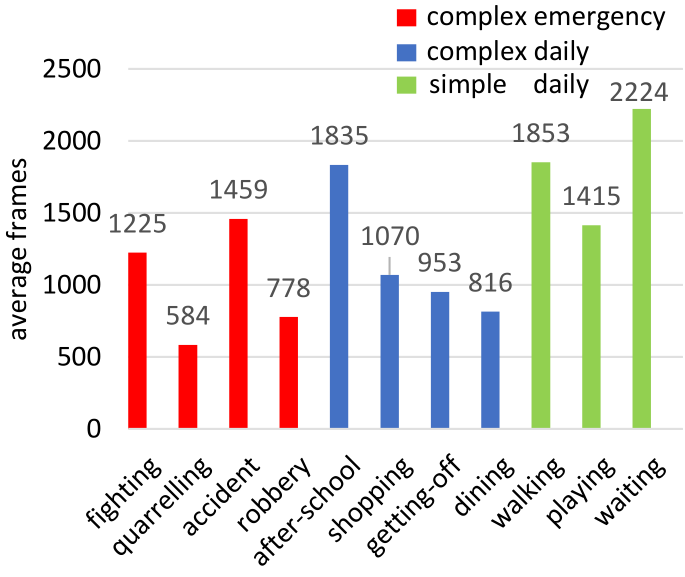}
		\caption{The distribution of events. Different colors represent different kinds of events.}
		\label{fig:sub-events_distribution}
	\end{minipage}
	\hspace{0.5cm}
	\begin{minipage}[t]{0.29\linewidth}
		\centering
		\includegraphics[width=\textwidth]{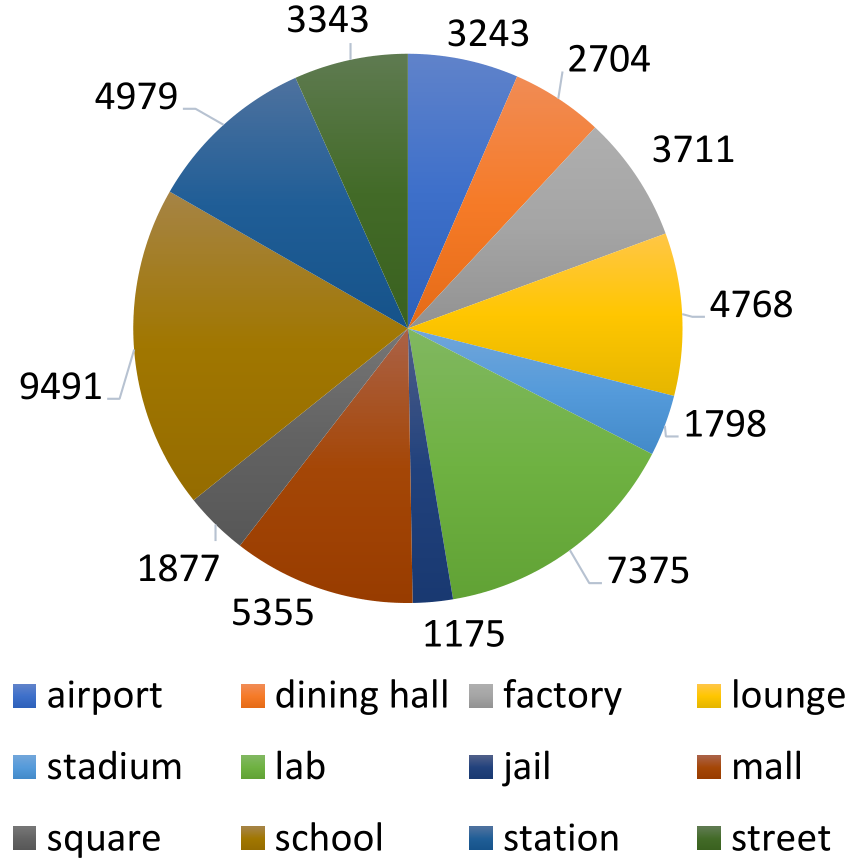}
		\caption{The frame number distribution of different scenes in HiEve dataset.}
		\label{fig:scean_distribution}
	\end{minipage}
\end{figure*}

\noindent
\textbf{Collection} We start by selecting several crowded places with complex and diverse events for video collection. The videos are collected from two sources. The first part of the videos was obtained by ourselves where the consents of participants were obtained in advance. The second part of the videos was collected from online repositories such as YouTube. We include them in our dataset according to the guidance of \textit{Fair use on YouTube}. We also have verified that all personally identifiable information (e.g., faces) was blurred in these videos and cannot be used to identify a specific subject. Note that the video collection process (including participants recruitment, video shooting, and online video collection) follows the guidance of the IRB review of our institute, which ensures HiEve doesn't violate individual privacy or other legal or ethical standards.  
In total, our video sequences are collected from 12 different scenes: \textit{airport, dining hall, factory, lounge, stadium, jail, mall, square, school, station and street}.  \autoref{fig:scean_distribution} shows the frame number of different scenes in HiEve.
As illustrated in the workflow in \autoref{fig:collection}, for each scene, we keep several videos captured at different sites and with different types of events happening to ensure the diversity of scenarios. Moreover, data redundancy is avoided through manual checking. 
Finally, 32 real-world video sequences in different scenes are collected (with 10 videos obtained by ourselves and 22 videos collected from online repositories), each containing one or more complex events. These video sequences are split into training and testing sets of 19 and 13 videos. Both our own collected videos and online resources videos have a roughly sixty-forty split in training and testing. 
The detailed training-setting split as well as the detailed information of each video (including FPS, resolution, frame number, and source) can be found  in \url{http://humaninevents.org/data.html}

\noindent
\textbf{Annotation} We manually annotated the HiEve dataset by cooperating with a professional annotation company, which owns experienced data annotators and has provided annotation services to many well-known benchmarks. All the data are labeled under a standard procedure to ensure their quality. The annotation procedure is as follows: 
\textbf{\textit{First}}, we annotate poses for each person in the entire video. Different from PoseTrack and COCO, our annotated pose for each body contains 14 keypoints (\autoref{fig:keypoints}): \textit{nose, chest, shoulders, elbows, wrists, hips, knees, ankles}.  Specially, we skip pose annotation which falls into any of the following conditions: (1) heavy occlusion (2) area of the bounding box is less than 500 pixels. \autoref{fig:pose_sample} presents some pose and bounding-box annotation examples.
\textbf{\textit{Second}}, we annotate actions of all individuals every 20 frames in a video. For group actions, we assign the action label to each group member participating in this group activity. In total, we defined 14 action categories: \textit{walking-alone, walking-together, running-alone, running-together, riding, sitting-talking, sitting-alone, queuing, standing-alone, gathering, fighting, fall-over, walking-up-down-stairs, crouching-bowing}. \autoref{fig:action_sample} shows some samples of different actions in HiEve. \textit{\textbf{Third}}, In order to guarantee the quality of our annotation results, we also conduct a temporally sequential annotation process. Specifically, we inherit all annotations from the previous frame and then update the annotations according to object appearances in the current frame. This process can both maintain high temporal consistency in the annotation results and greatly reduce the annotation burden at the same time. \textit{\textbf{Finally}}, all annotations are double-checked to ensure their quality. Specifically, there are two groups of humans to conduct data annotation. All videos are first sent to one group for the 1st round of labeling following the standard annotation procedure. After the 1st round of annotation, the labeled data are then sent to another group for double-checking. During the double-check, annotations of each sample will be evaluated with a confidence score (value from 0 to 10), which indicates the confidence of labeling. Then, data with less than 9 confidence scores will be sent back to the forehead group for the 2nd round annotation. We repeat the above process until all annotations satisfy the rule of confidence score. Moreover, we set a maximum iterations (iter=4 in our annotation) for correcting the annotation cross-check process.

It should be noted that in order to maintain the completeness and consistency in the annotation results for all objects in a scene, we annotate both visible and invisible keypoints \& bounding boxes.  For invisible keypoints and boxes, we infer their location from the motion cues from previous frames or by observations, and assign them with an additional `invisible' label.  However,  during the performance evaluation stage, we only evaluate performances based on the visible keypoints \& bounding boxes, while the `invisible' keypoints \& bounding boxes are not included.  This can make our evaluation results more accurate and reliable.

\begin{figure*}[ht]
	\centering
	\begin{minipage}[t]{0.24\linewidth}
		\centering
		\includegraphics[width=0.9\textwidth,height=0.80\textwidth]{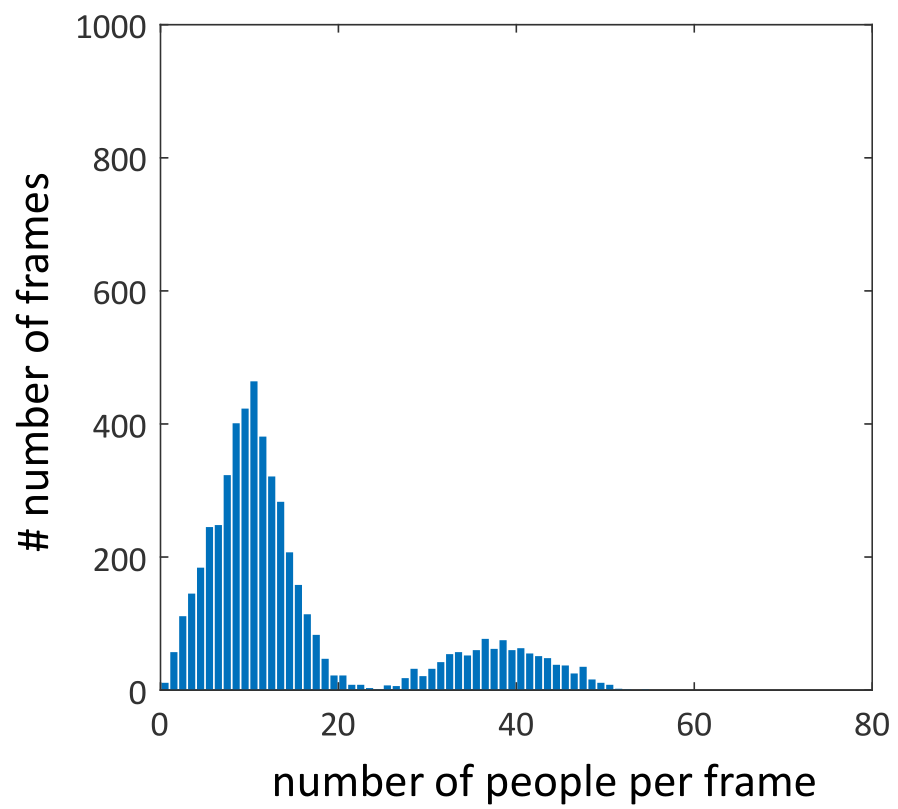}
		\subcaption{MOT17}
	\end{minipage}  
	\begin{minipage}[t]{0.24\linewidth}
		\centering
		\includegraphics[width=0.9\textwidth]{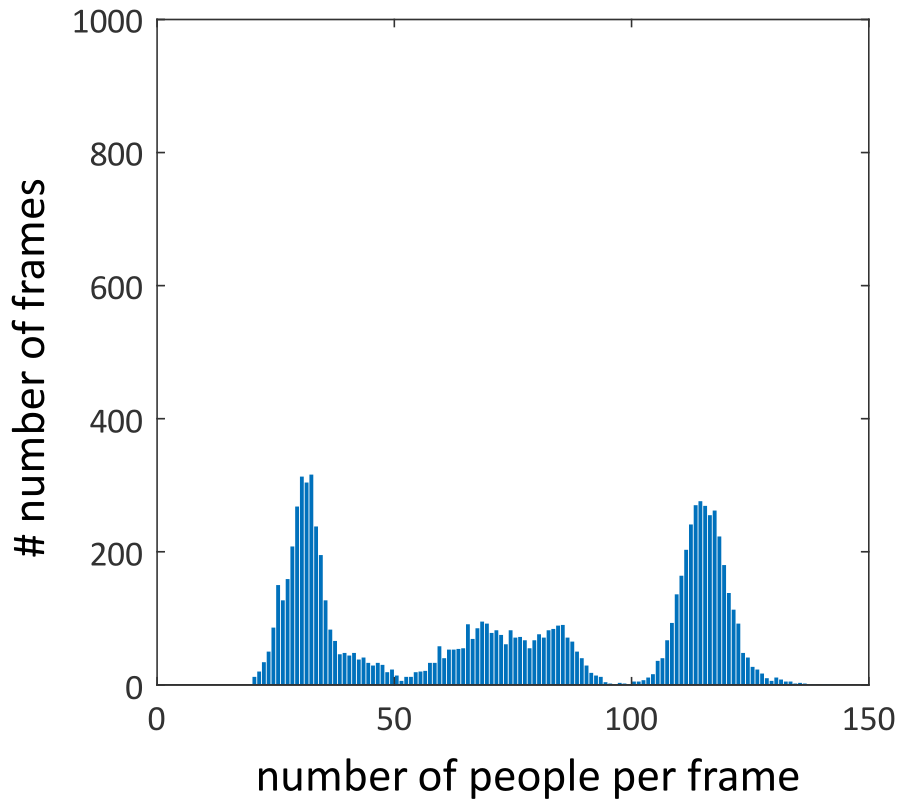}
		\subcaption{MOT20}
	\end{minipage} 
	\begin{minipage}[t]{0.24\linewidth}
		\centering
		\includegraphics[width=0.9\textwidth]{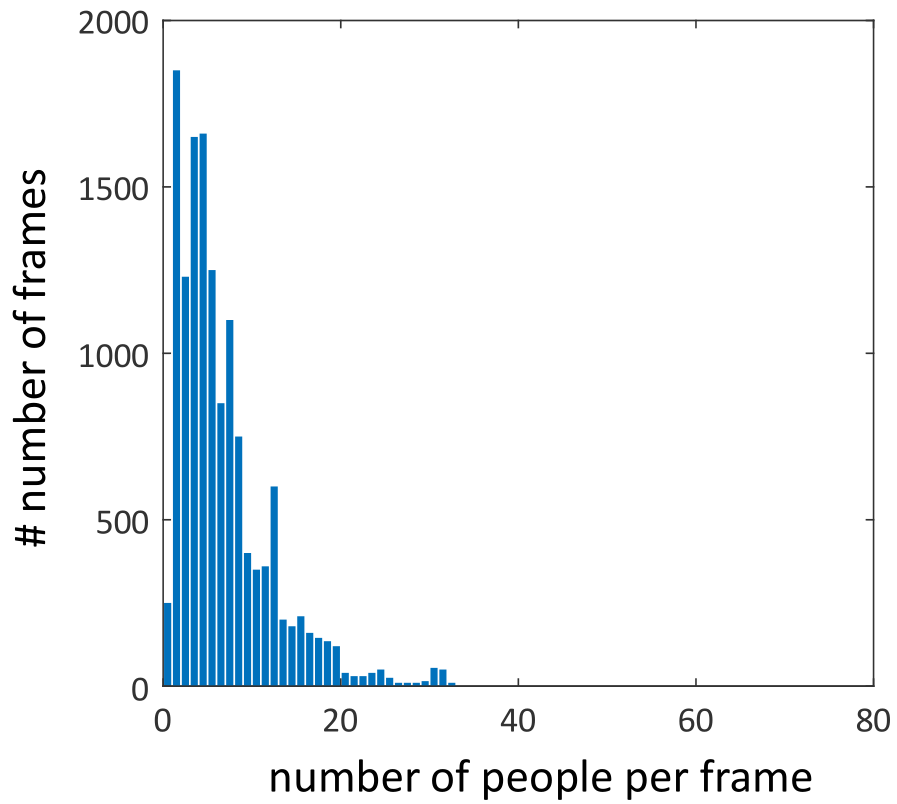}
		\subcaption{PoseTrack}
	\end{minipage} 
	\begin{minipage}[t]{0.24\linewidth}
		\centering
		\includegraphics[width=0.9\textwidth]{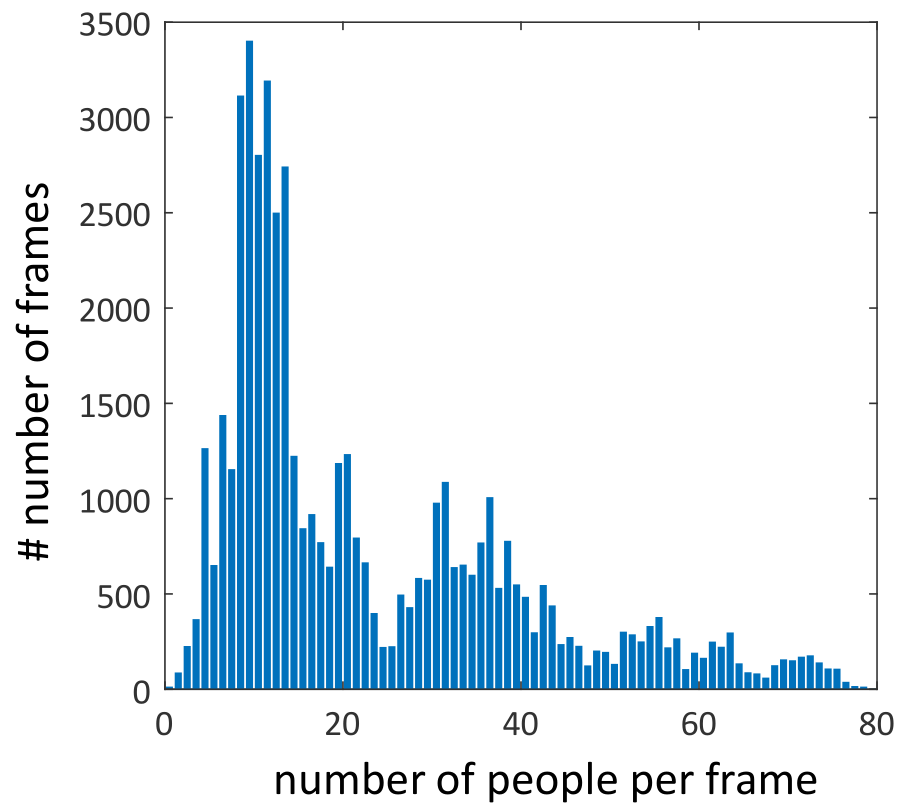}
		\subcaption{HiEve}
	\end{minipage} 
	\caption{The distribution of the number of people per frame in MOT17, MOT20, PoseTrack and HiEve dataset. The scenes in HiEve dataset owns more people.}
	\label{fig:person_number}
\end{figure*}

\subsection{HiEve Statistics}

Our dataset contains 32 video sequences mostly longer than 900 frames. Their total length is 33 minutes and 18 seconds. 
\autoref{table:comprision} shows the basic statistics of our HiEve dataset: It contains 49,820 frames, a record number of poses (1,099,357), the largest number of action instances (56,643) under complex events, as well as one of the largest numbers of trajectories (2,687) lasting for longer time (with an average trajectory length of 485 frames).
\begin{figure*}[ht]
	\centering
	\begin{minipage}[t]{0.2\linewidth}
		\vspace{0.5cm}
		\includegraphics[width=0.7\textwidth]{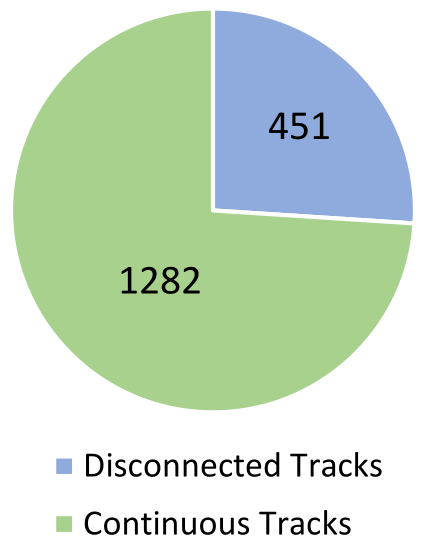}
		\caption{Number of disconnected and continuous tracks in training set.}
		\label{fig:disconnected_tracks}
	\end{minipage}
	\hspace{1pt}
	{\vrule width0.6pt}
	\hspace{1pt}
	\begin{minipage}[t]{0.75\linewidth}
		\vspace{0.5cm}
		\hspace{0.5cm}
		\begin{minipage}[t]{0.21\linewidth}
			\centering
			\includegraphics[width=1\textwidth,height=0.90\textwidth]{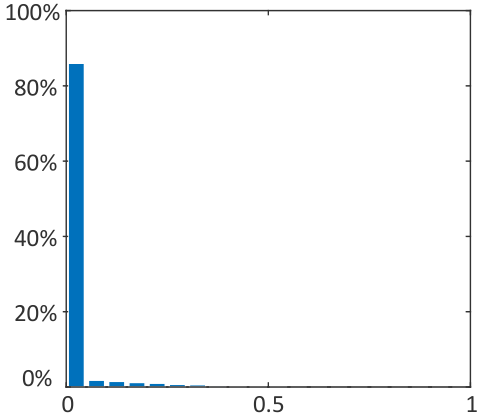}
			\subcaption{MPII}
		\end{minipage}  
		\begin{minipage}[t]{0.21\linewidth}
			\centering
			\includegraphics[width=1\textwidth]{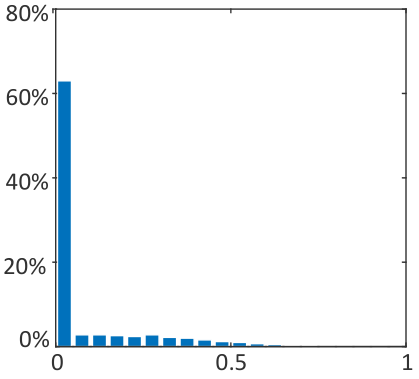}
			\subcaption{MSCOCO}
		\end{minipage}
		\begin{minipage}[t]{0.21\linewidth}
			\centering
			\includegraphics[width=1\textwidth]{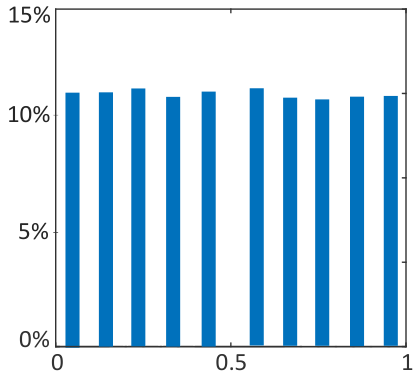}
			\subcaption{CrowdPose}
		\end{minipage} 
		\begin{minipage}[t]{0.21\linewidth}
			\centering
			\includegraphics[width=1\textwidth]{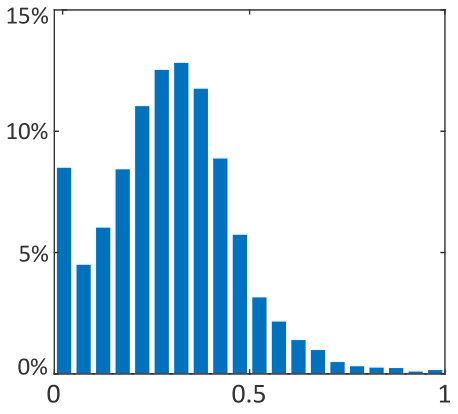}
			\subcaption{HiEve}
		\end{minipage} 
		\vspace{0.2cm}
		\caption{\textit{CrowdIndex} distributions of MPII, MSCOCO, CrowdPose, and our HiEve dataset. MSCOCO is dominated by uncrowded images. while HiEve dataset pays more attention on crowded cases.}
		\label{fig:crowdindex}
	\end{minipage}
\end{figure*}
\begin{figure*}[ht]
	\centering
	\begin{minipage}[t]{0.4\linewidth}
		\includegraphics[width=0.9\textwidth]{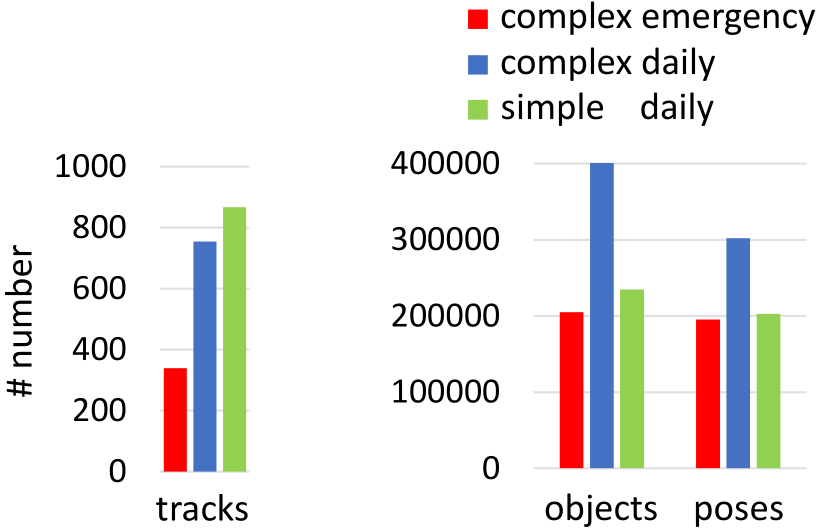}
		\vspace{2.5pt}
		\caption{The number of tracks, objects and poses in events. Different  colors  represent  different  kinds  of events.}
		\label{fig:pose_track_poses_num}
	\end{minipage}
	\hspace{2pt}
	{\vrule width0.6pt}
	\hspace{2pt}
	\begin{minipage}[t]{0.5\linewidth}
		\vspace{-4cm}
		\begin{minipage}[t]{0.45\textwidth}
			\centering
			\includegraphics[width=1\textwidth,height=0.90\textwidth]{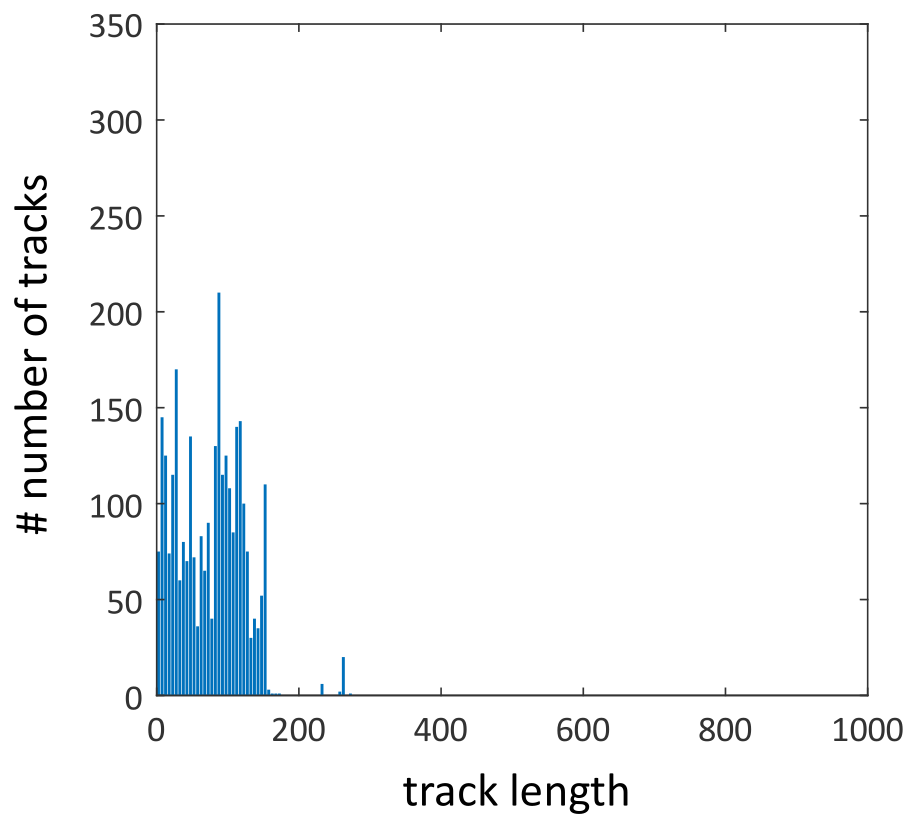}
			\subcaption{PoseTrack}
		\end{minipage}  
		\begin{minipage}[t]{0.45\textwidth}
			\centering
			\includegraphics[width=1\textwidth]{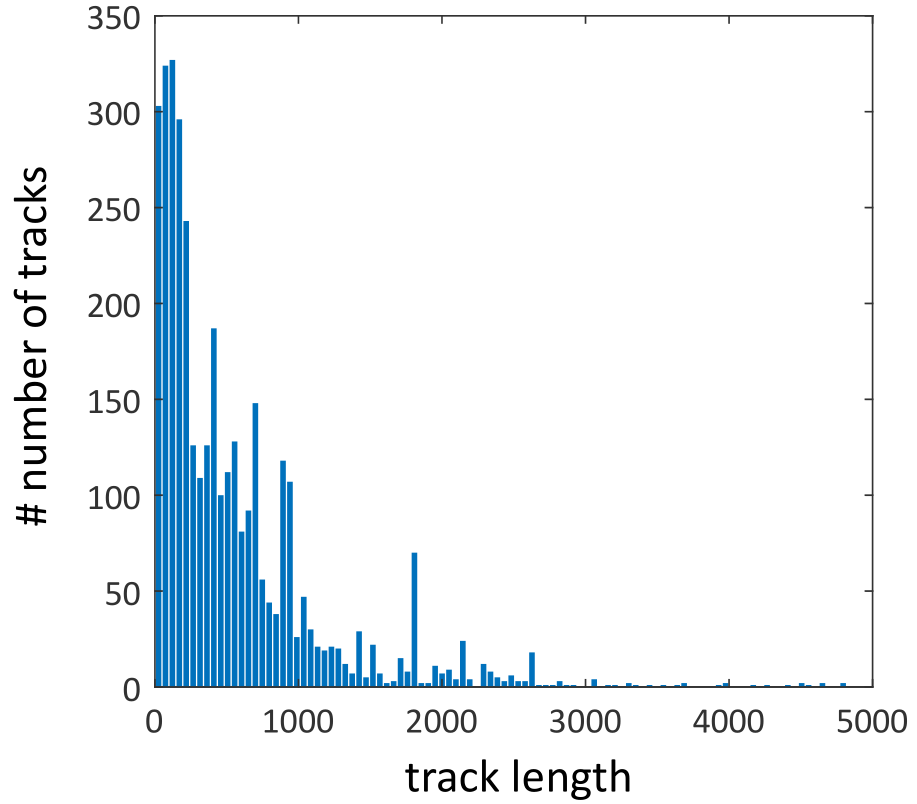}
			\subcaption{HiEve}
		\end{minipage} 
		\caption{The distribution of the length of track in PoseTrack and HiEve dataset.}
		\label{fig:track_length}
	\end{minipage}
\end{figure*}


To further illustrate the characteristics of our dataset, we conduct the following statistical analysis.

\textbf{\textit{First}}, we analysis some statistic information across different events. In terms of video content, we could group our video sequences into 11 \textbf{events}: \textit{fighting, quarreling, accident, robbery, after-school, shopping, getting-off, dining, walking, playing and waiting.} Each event contains different amount of participants and action types. Then, according to the complexity of these events, we further grouped these events into 3 categories: \textbf{\textit{complex emergency event, complex daily event, and simple daily event}}. In this way, we can construct the relationship between action, event, and  category with a bottom-up manner, where each event may contain multiple actions, and each event category includes multiple event types (cf. \autoref{table:sub-events}). This hierarchical structure also allows for better statistical analysis of our HiEve dataset. We first present the number of poses, objects, and tracks for the above 3 events in \autoref{fig:pose_track_poses_num}. From this figure, we can see that (1) In our HiEve dataset, complex events (including complex emergency and complex daily) contain more human-centric instances (i.e., tracks, objects, and poses) compared to simple events.  (2) Among the three event categories, complex daily events exhibit the largest number of poses and objects.  Meanwhile, complex emergency events also have a considerable number of poses and objects compared to simple daily events. Moreover, \autoref{fig:sub-events_distribution} presents the average frame number of each event category. It can be seen that both the complex daily event and complex emergency event contain a considerable number of video frames in our HiEve dataset, which further indicates that our dataset is dominated by complex events. All these observations demonstrate the significant challenges posed by our dataset.

\textbf{\textit{Second}}, we present the number of people per frame in our dataset in \autoref{fig:person_number} demonstrating that the scenes in our video sequence have more people than MOT17 and PoseTrack~\cite{posetrack2018}, making our tracking task more difficult. Although MOT-20~\cite{mot2020} collects some video sequences with more people (up to 141 people), it only covers limited scenarios and human actions. 

\textbf{\textit{Third}}, we adopt the \textit{Crowd Index} defined in Crowdpose~\cite{crowdpose2019} to measure the crowding level of our dataset. For a given frame, its Crowd Index(CI) is computed as:

\begin{equation}
	CI = \frac{1}{n}\sum_{i=1}^{n}\frac{N_i^b}{N_i^a}  
	\label{eq:crowdindex} 
\end{equation}
where $n$ is the total number of persons in this frame. $N_{i}^a$ denotes the number of joints from the $i^{th}$ human instance and $N_i^b$ is the number of joints located in bounding-box of the $i^{th}$ human instance but not belonging to the $i^{th}$ person. We evaluate the \textit{Crowd Index} distributions of our HiEve dataset and the pose dataset MSCOCO~\cite{coco}, MPII~\cite{mpii2014}, and CrowdPose~\cite{crowdpose2019}.  \autoref{fig:crowdindex} shows that our HiEve dataset pays more attention to crowded scenes while other benchmarks are dominated by uncrowded ones. This characteristic enables our HiEve to comprehensively evaluate various pose estimation methods, ranging from simple cases to hard crowded scenes. Moreover, we need to clarify that the CrowdPose dataset is carefully selected from three public datasets (MSCOCO, MPII, and AI Challenge) according to the CrowdIndex. In this way, it has a near-uniform distribution of CrowdIndex. On the contrary, our HiEve dataset is a newly collected large-scale dataset rather than a selected subset of available benchmarks.
\begin{figure*}[t]
	\centering
	\begin{minipage}[t]{0.45\textwidth}
		\centering
		\includegraphics[width=\linewidth]{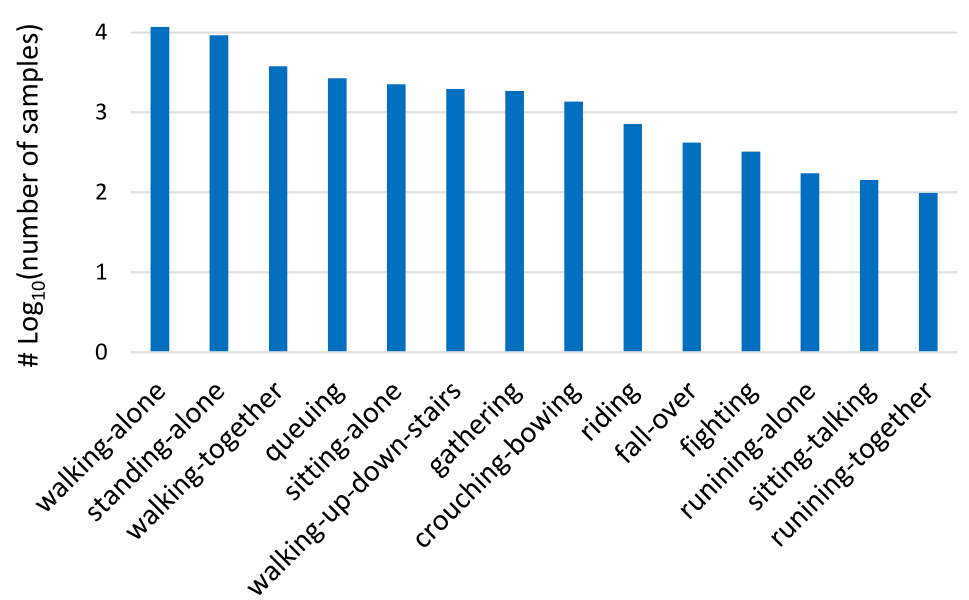}
		\caption{Sample distribution of all action classes in the HiEve dataset. Note that present the $log_{10}$ of number of samples for a better visualization.}
		\label{fig:action_distribution}
	\end{minipage}
	\hspace{5pt}
	\begin{minipage}[t]{0.45\textwidth}
		\centering
		\includegraphics[width=\linewidth]{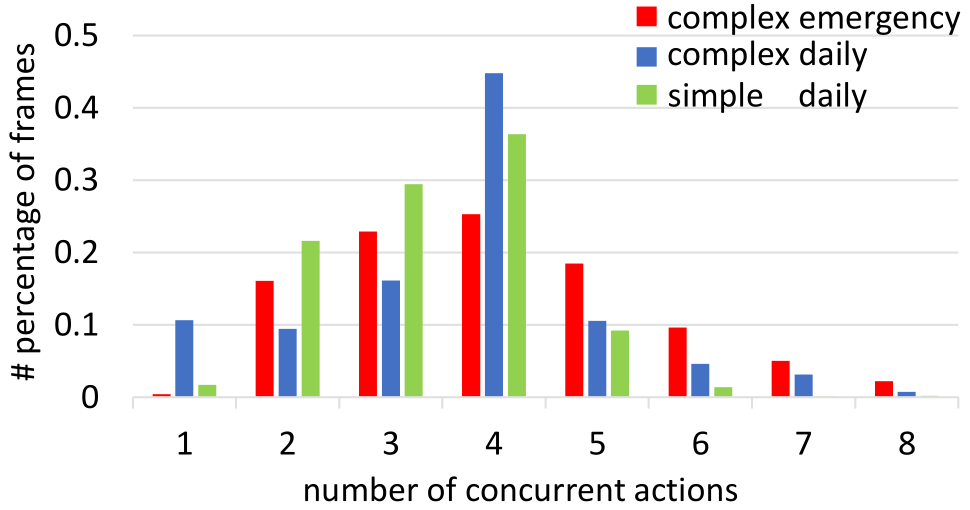}
		\caption{The distribution of the number of concurrent action in HiEve dataset. Different colors represent different kinds of events.}
		\label{fig:concurrent}
	\end{minipage}
	\label{fig:distribution}
\end{figure*}

\begin{figure}[t]
	\centering
	\includegraphics[width=0.9\linewidth]{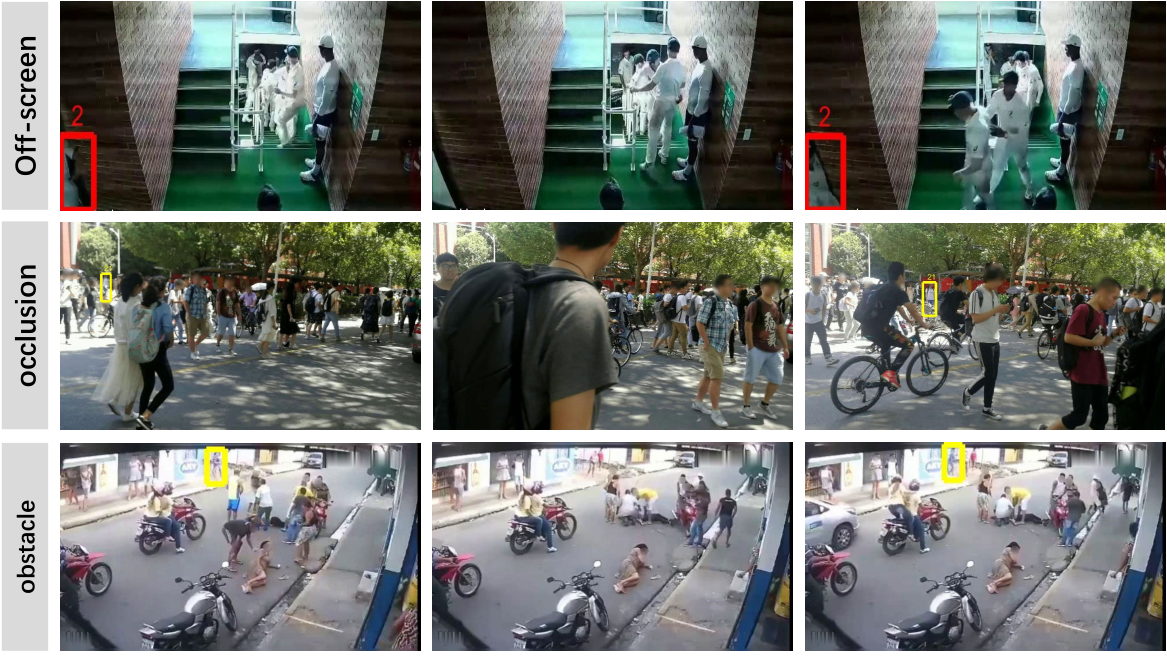}
	\caption{Examples of disconnected tracks (highlighted with bounding box)}\label{fig:example_disconnect_tracks}
\end{figure}

\textbf{\textit{Fourth}}, we analyze the ratio of disconnected human tracks in our dataset. \textit{Disconnected human tracks} are defined as trajectory annotations where the bounding boxes are not available on some frames due to: (1) One object temporally moves out of the camera view and moves back sometime later. (2) One object is severely occluded by foreground objects or certain obstacles for a long time so that annotators can not assign an approximate bounding box to it (as exemplified in \autoref{fig:example_disconnect_tracks}). It is noticeable that in datasets like PoseTrack~\cite{posetrack2018}, the reappearance of one individual in the scene is considered as the start of a new trajectory instead of the continuation of the original track before disappearing, in this manner these datasets will contain more tracks with shorter endurance (as reflected in \autoref{fig:track_length}). In contrast, in HiEve we assign the tracks before and after disappearing with the same ID, so as to encourage algorithms which can properly handle long-term re-identification. The numbers of disconnected and continuous tracks in the training set are reported in \autoref{fig:disconnected_tracks}. The statistical results show that the proportion of disconnected tracks is non-negligible supporting algorithms which could handle complex cases and crowded scenes.

\textbf{\textit{Finally}}, the distribution of all action classes in our dataset is shown in \autoref{fig:action_distribution} and could be regarded as a long-tailed sample distribution. \autoref{fig:concurrent} demonstrates the complex events in our dataset have more concurrent events, which means that the complexity and difficulty of identifying behaviors in such scenes will increase.

Overall, these statistics further prove that HiEve is a large-scale and challenging dataset dominated by complex events.

\section{Task and Metric}
\label{section:metric}
With the collected video data and available annotations, HiEve poses four tasks for the evaluation of video analysis algorithms. For each task, we adopt some widely used metrics. Meanwhile we also design some \textbf{new metrics} to measure the performance on crowded and complex scenes.
\subsection{Multi-person tracking}
This task is proposed to estimate the location and corresponding trajectory of each identity throughout a video. Traditional metrics MOTA, MOTP~\cite{mot16}, ID F1 Score, ID Sw~\cite{idf2016}, and ID Sw-DT are selected to perform evaluation. Apart from these traditional metrics, our HiEve dataset also includes the novel HOTA~\cite{hota} (Higher Order Tracking Accuracy) metric for evaluating MOT performance. HOTA is a comprehensive metric that considers various aspects of multi-object tracking, such as detection, localization, identity preservation, and temporal consistency. We believe that the incorporation of these metrics will provide a more accurate and reliable evaluation of tracking algorithms on our dataset.

Besides, in order to evaluate how algorithms perform on tracks with disconnected parts, we design a \textbf{weighted MOTA (w-MOTA) } metric. This metric is computed in a similar manner as MOTA except that we assign a higher weight $\gamma$ to the ID switch cases happening in disconnected tracks, consequently the metric can be formulated as
	$$
	\textbf{w-MOTA} = 1-(N_{fp} + N_{fn} + N_{sw} + (\gamma-1)N_{sw-dt}) / N_{gt}
	$$
	where $N_{fp}$ and $N_{fn}$ are the number of false positive and false negative, $N_{sw}$ is the total times of ID switch, $N_{sw-dt}$ is the ID switch times happening in disconnected tracks and $N_{gt}$ is the number of bounding boxes in annotations.

\subsection{Multi-person pose estimation}
This task aims to estimate specific keypoints on human skeleton. Compared with MPII Pose and MSCOCO Keypoints, our dataset involves more real-scene pose patterns in various complex events. We adopt \textit{Average Precision} (\textbf{AP@$\alpha$}) for measuring multi-person pose accuracy. The evaluation protocol is similar to DeepCut~\cite{deepcut2016}, if a pose prediction has the highest PCKh~\cite{mpii2014} with a certain ground-truth, then it can be assigned to the ground truth. Unassigned predictions are counted as false positives.  
$\alpha$ is the specific distance threshold for computing PCKh. We take the average value of AP@0.5, AP@0.75, and AP@0.9 as an overall measurement AP@avg.

To further avoid the methods only focusing on simple cases or uncrowded scenarios in the dataset (although \autoref{fig:crowdindex} has shown that our dataset contains a large number of crowded and complex scenarios), we will assign larger weights to a test image during evaluation if it owns: (1) higher \textit{Crowd Index} (2) anomalous behavior (e.g. fighting, fall-over, crouching-bowing). To be specific, the weights for the $t^{th}$ frame in one video sequence can be formulated as:
	$$
	w^P_t = c_{1}e^{CI_t} + c_{2}N_{t}
	$$
	where $CI_{t}$ is the crowd index on $t^{th}$ frame calculated via \autoref{eq:crowdindex}, $N_t$ denotes the number of categories of anomalous actions. During our evaluation, the coefficients $c_{1},c_{2}$ are set to $2,1$ respectively.
The values of AP calculated with assigned weights are called \textbf{weighted AP (w-AP)}. Besides, we calculate  \textbf{w-AP@avg} in the similar way with AP@avg.

\begin{figure*}[ht]
	\centering
	\begin{minipage}[t]{0.28\textwidth}
		\centering
		\includegraphics[width=0.85\linewidth,height=0.95\linewidth]{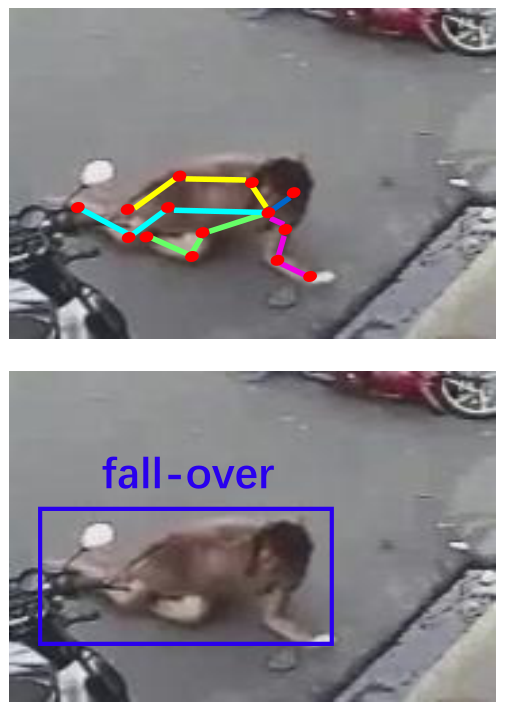}
		\caption{The keypoints distribution may indicate the `fall-over'.}
		\label{fig:pose2action}
	\end{minipage}
	\begin{minipage}[t]{0.7\textwidth}
		\centering
		\includegraphics[width=\linewidth]{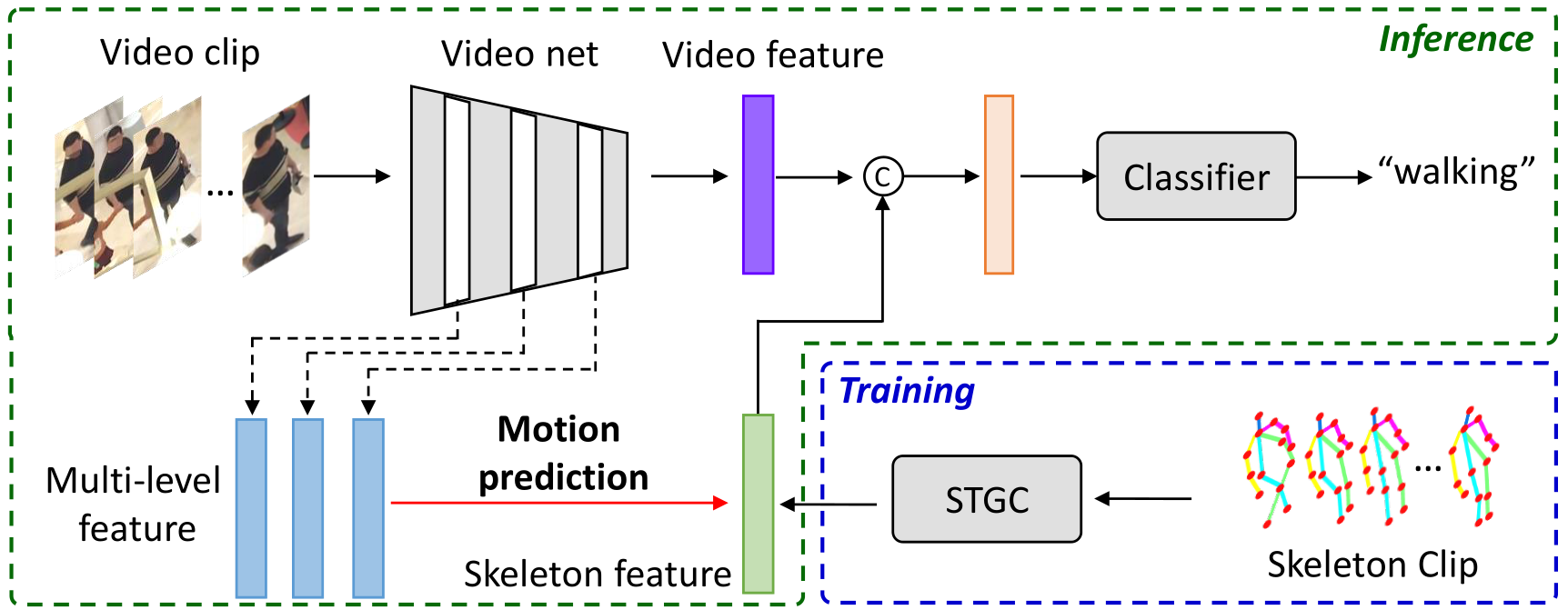}
		\caption{The framework of pose-aware action recognition enhanced baseline.}
		\label{fig:pose_aware}
	\end{minipage}
	\vspace{-6pt}
	\begin{minipage}[t]{0.95\textwidth}
		\centering
		\includegraphics[width=\linewidth]{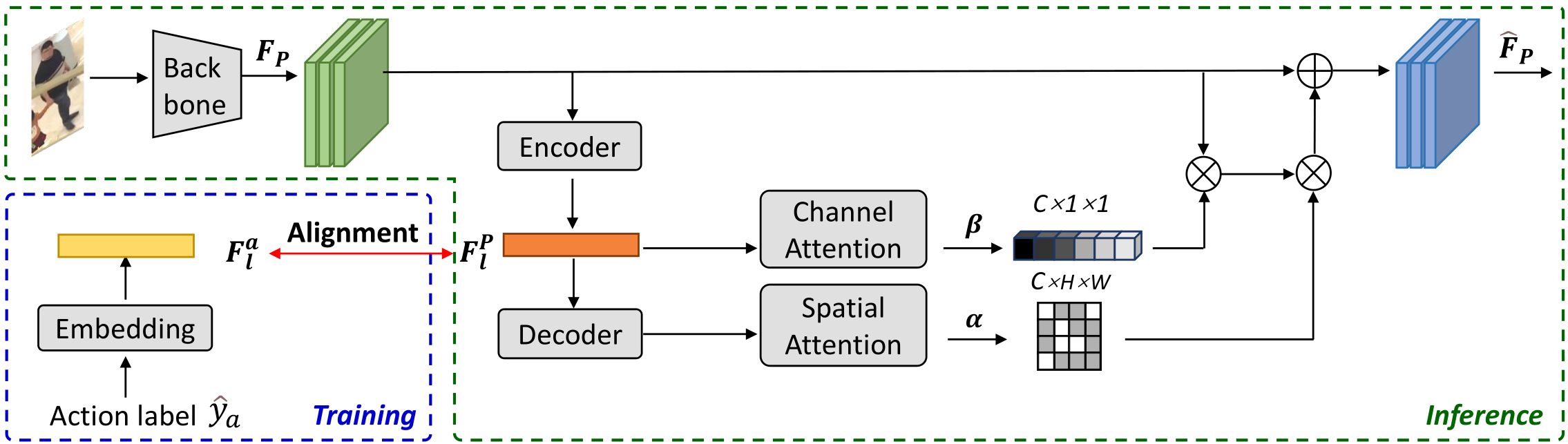}
		\caption{The framework of action-guided pose estimation enhanced baseline.}
		\label{fig:action_embedding}
	\end{minipage}
\end{figure*}

\subsection{Pose tracking}
This task requires to provide temporally consistent poses for all people visible in the videos. Compared with PoseTrack, our dataset is much larger in scale and includes more frequent occlusions. Evaluation metrics MOTA and MOTP are also adopted in this task.
\subsection{Action recognition}
The action recognition task requires participants to simultaneously detect specific individuals and assign correct action labels to it on every sampled frame. Compared with AVA challenge~\cite{Gu2018AVAAV}, our action recognition track does not only contain atomic level action definition but also involves more interactions and occlusion among individuals, making recognition more difficult.
We adopt the frame mAP (f-mAP@$\alpha$), which is widely used to evaluate spatial action detection accuracy on a single frame, as the basic metric in this task. $\alpha$ is the specific IOU threshold to determine true/false positive. We report the mean value of f-mAP@0.5, f-mAP@0.6, and f-mAP@0.75 as an overall measurement of f-mAP, we denote this measurement as f-mAP@avg.

Furthermore, considering the unbalanced distribution of the action categories in the data set, it is appropriate to assign smaller weights to the test samples belonging to dominated categories. In addition, we assign a larger weight to frames under crowded and occluded scenarios to encourage models to perform better in complex scenes.  
The frame mAP value calculated with these assigned weights is called \textbf{weighted frame-mAP (wf-mAP)}. Similarly to f-mAP@avg, we also report \textbf{wf-mAP@avg} as an overall measurement of wf-mAP.

\section{Enhanced baselines with cross-annotation}
The main advantage of HiEve is that it provides a wide range of human-centric annotations (tracking, pose, action), thus encouraging researchers to design visual algorithms by utilizing annotations from different types and aspect. This results in more comprehensive and accurate human-centric visual analysis system. To validate the above ability of HiEve, we design two simple baselines for action recognition and pose estimation tasks based on HiEve in this section.

\subsection{Pose-aware action recognition}

Skeleton-based action recognition~\cite{skeleton1,skeleton2,skeleton3} has attracted much attention due to its innate ability to represent motion. Current skeleton-based algorithms are predominantly developed and evaluated using benchmarks with simple scenes, such as the NTU-RGB-D~\cite{ntu}, which comprises only one or two individuals per frame. However, achieving accurate pose estimation in complex scenarios, particularly those with heavy occlusion, proves exceedingly difficult, limiting the application of skeleton-based methods.  Therefore, the potential of skeleton representation under complex scenes for action recognition still remains under exploration. Leveraging the diverse annotations in HiEve, we establish an enhanced baseline for RGB-based action recognition, where skeleton information is implicitly learned and integrated into the video representation. Its overall architecture is illustrated in \autoref{fig:pose_aware}. It is worth noting that, unlike traditional skeleton-based approaches, we don't require human poses during inference. Compared to RGB-based methods, the only additional information we employ is the pose annotation of training data provided by HiEve. In summary, our proposed paradigm enables us to utilize pose information to facilitate action recognition while concurrently avoiding incorrect pose estimation under complex events.

\subsubsection{Multi-level motion prediction}
The skeleton sequence contains more pose motion patterns, whereas the video representation includes more appearance-related motion information. Based on the various annotation for training data in HiEve, we can leverage the pose annotation to facilitate the video feature learning by providing complementary pose-aware motion.

Given a video clip, the video-based pipeline extracts video features $f_{v}\in \mathbb{R}^d$ using a video-specific model (e.g., I3D~\cite{kinetics2017} , SlowFast~\cite{slowfast}). Meanwhile, we could resort to human pose annotation provided by HiEve to generate a corresponding skeleton graph sequence $G$ for this clip. Since the graph convolution network (GCN) has been widely used to process the skeleton sequences, we also resort to the GCN module proposed in STGCN~\cite{stgcn2018} to extract the skeleton feature.
	\begin{equation}
		f_p = \texttt{STGCN}(G)
	\end{equation}
	where $f_p\in \mathbb{R}^{d}$ indicates the skeleton graph feature output by GCN, which we can name pose-aware feature.

To empower the video network to obtain pose-aware motion by itself, we design a multi-level motion prediction task for the video stream. It encourages the video network to predict the pose-aware motion representation using multi-level video features. Meanwhile, we find it beneficial to predict the direction $f_p^c$ and length $\lVert f_p \Vert$ of $f_p$ separately. The $f_p$ vector can be decomposed into its direction and length, so we can re-write it as:
	\begin{equation}
		f_p = \frac{f_p}{\lVert f_p \Vert} \cdot \lVert f_p \Vert = f_p^c \cdot \lVert f_p \Vert
	\end{equation}
	The video features across layers in CNN models contain multi-level and multi-grained action patterns, so it's promising for them to learn a robust motion representation. Therefore, we use video features from multiple stages of the model to conduct this prediction. For each feature map $m_l\in \mathbb{R}^{d_l}$ output by the 3D CNN model in stage-$l$, we predict the corresponding pose-aware motion vector by linear transformation:
	\begin{gather}
		r_l^c = \frac{W_l^c m_l + b_l^c}{\lVert W_l^c m_l + b_l^c \Vert}, ~r_l^s = W_l^s m_l + b_l^s
	\end{gather}
	where $W_l^c \in \mathbb{R}^{d\times d_l}$ and $b_l^c$ are the parameters of direction prediction, while $W_s^c \in \mathbb{R}^{1\times d_l}$ and $b_l^s$ belong to the length prediction.
	We aggregate multiple predictions from multi-level features by: 
	\begin{gather}
		r=r^s\cdot r^c,~\text{where}~r^c = \frac{\sum_{l=1}^{L}r_l^c}{\lVert \sum_{l=1}^{L}r_l^c \Vert}, ~r^s = \sum_{l=1}^{L}r_l^s 
	\end{gather}
	Moreover, we add a prediction loss term to encourage the predicted motion vector $r$ to be close to the $f_p$:
	\begin{gather}
		\mathcal{L}_{pred} = \lVert f_p^c - r^c \Vert^2_2 + (r^s - \lVert f_p\Vert)^2
	\end{gather}
	Finally, the predicted feature vector is concatenated with the video feature $f_v$, which provides the video feature with complementary pose-specific motion patterns.

\subsubsection{Implementation Details}
The Slowfast-ResNet50~\cite{slowfast} is chosen as our backbone for video feature extraction. Moreover, we follow the official setting of SlowFast to keep the same temporal resolution at different stages of ResNet. Regarding the additional overhead introduced by our baseline, it only adds approximately 25\% GFLOPs to the vanilla SlowFast (an increase from 65.7 GFLOPs to 84.6 GFLOPs). Faster-RCNN detector is used to detect persons during testing. $L=3$ in our default setting and the feature maps output by stage-1, 2, 3 are globally pooled to form as the multi-level feature $m_1, m_2, m_3$. the final feature dimension $d=2034$. We uniformly sample 16 frames for each video and each input frame is cropped into $256\times 256$ during training and inference. 
	The total loss for training is defined as:
	\begin{equation}
		\mathcal{L} = \mathcal{L}_{cls} + \mathcal{L}_{pred}
	\end{equation}
where the $\mathcal{L}_{cls}$ is the cross-entropy loss adopted in classification task. During inference, since the pose annotation is not available, we straightly use the predicted pose-aware motion feature as the input for classifier.

\subsection{Action-guided pose estimation}

Although skeleton-based action recognition has been well developed, only a few methods~\cite{posewithaction2015,posewithaction2017} paid attention to its reverse paradigm, i.e., how action prior can help pose estimation. Luckily, thanks to the diverse annotations of HiEve, we build a simple yet effective baseline method for pose estimation, which enhances the pose learning stream by prior knowledge of action. As shown in \autoref{fig:action_embedding}, the algorithm mainly comprises two modules: action-guided domain alignment module (ADAM) and pose refinement module (PRM) module, where ADAM aligns the feature representation between the domain of action and pose, while PRM utilizes the aligned feature to refine the pose estimation results. Compared to previous approaches that attempt to leverage the action knowledge to facilitate the pose estimation, our method offers several advantages: First, it is free from utilizing additional action predictors during inference, which is necessary for most previous methods~\cite{posewithaction2015,posewithaction2017}. Second, we only added negligible overhead to the pose estimation stream. Thirdly, our method can be easily extended to most current pose estimation algorithms. It's worth noting that some approaches integrate pose and action learning into a multi-task learning framework~\cite{multi-task} or a unified model. Different from them, our focus remains on the pose estimation task.

\subsubsection{Action-guided domain alignment}
Some special location relationships between human keypoints tend to indicate a certain anomalous behavior. For example, as illustrated in \autoref{fig:pose2action}, a human skeleton yielding a dense and horizontal keypoints distribution is usually associated with the \textit{`fall-over'} action. Vice versa, the action category can provide reliable prior knowledge on keypoints location. Moreover, the incorrect keypoints location could be revised by these knowledge. With this observation, we propose an action-guided domain alignment module (ADAM), where we regard the pose and action as information from two different domains. The ADAM aims at building a mapping between them, such that the two domains are close in feature space.

Follow the framework of top-down pose estimation, the pose feature $\mathbf{F}_{p}$ of single person is extracted by a base convolution network. Then, an encoder $\mathbf{E}$ with a series of down-sample operations squeezes the pose feature into a latent feature $\mathbf{f}_{l}^{p}\in \mathbb{R}^{d}$. To extract action information, we embed the one-hot action label vector $\mathbf{\hat{y}}_{a}$ of this person into a latent feature $\mathbf{f}_{l}^{a}\in \mathbb{R}^{d}$ through a linear transformation $\mathbf{T}$. The above process could be formulated as:
\begin{align*}
	\mathbf{f}_{l}^{p} = \mathbf{E}(\mathbf{F}_{p}), ~
	\mathbf{f}_{l}^{a}=\mathbf{T}(\mathbf{\hat{y}}_{a}),~ \mathbf{f}_{l}^{p},\mathbf{f}_{l}^{a} \in \mathbb{R}^{d} 
\end{align*}
Then, an alignment loss is calculated between latent features from two domains, which encourages feature consistency between them by minimizing their distance in the latent space:
\begin{equation}
	\mathcal{L}_{align} = \text{MSE}(\mathbf{f}_{l}^{p}, \mathbf{f}_{l}^{a}) 
	\label{eq:align}
\end{equation}
However, there exists some variance among human poses even though they belong to the same action category. Aligning all of them to the same action embedding is not ideal. Moreover, for each individual in a complex event, action spatial-context (e.g., group activity, occlusion, or interaction with neighbors) also affects its human pose. Therefore, apart from input individual $o_{n}$ itself, we also consider action information from its neighboring area $U(o_n)$ and person  $o_m, m=1,2,\dots,|U(o_n)|, o_m\in U(o_n)$ in this area. Then, we can utilize the self-attention mechanism~\cite{transformer2017} to get an instance-specific action embedding by aggregating the spatial-context action information.

Specifically, we first embed their relative geo-position as:
\begin{gather}
	\footnotesize
	d^{mn} = \left( \frac{|x_{m}-x_{n}|}{w_{n}},  \frac{|y_{m}-y_{n}|}{h_{n}}\right)^{T},~g^{mn} = \mathcal{E}_{P}(d^{mn})
\end{gather}
where $\mathcal{E}_{P}$ is positional encoding operation proposed in Transformer~\cite{transformer2017}, $x,y,w,h$ are the center coordinates, width, and height of person bounding box.
Then, combining the action category embedding with relative-position embedding, we calculate spatial-context action correlations as:
\begin{gather}
	\footnotesize
	\omega^{m n}=\frac{\left\langle W_{K} ((\mathbf{f}_{l}^{a})_m +g^{mn}), W_{Q} ((\mathbf{f}_{l}^{a})_n + g^{nn})\right\rangle}{\sqrt{d_{k}}}
\end{gather}
where $W_K, W_Q \in \mathbb{R}^{d_k \times d}$ are projection matrices.
Specially, we only consider people $o_m$ who satisfy $|d^{mn}|^{2}\leq 4.5$ in the neighboring area $U(o_n)$. The spatial-context are aggregated into the individual action information embedding in a residual sum manner as:
\begin{gather}
	\footnotesize
	\mathbf{f}^a_l = \mathbf{f}^a_l + \sum_{m\in U(n)} \omega^{m n} \cdot\left(W_{V} \cdot (\mathbf{f}_{l}^{a})_m\right),
\end{gather}
where $W_{V}$ is projection matrix. The updated action embedding $\mathbf{f}_{l}^{a}$ is finally provided for $\mathbf{f}_{l}^{p}$ to perform alignment (\autoref{eq:align}).

\subsubsection{Pose refinement}
To further improve the quality of pose estimation, we design a refinement module based on the latent pose features, which comprises two head structures: spatial refinement head (SR) and channel-wise refinement head (CR).

In pose estimation, the position of keypoints is reflected by the local responses in the spatial feature maps. Therefore, the SR intends to re-weight the spatial feature map by emphasizing specific skeleton position and suppressing inaccurate keypoints response. 
Corresponding to the encoder in ADAM, the SR applies an decoder, which consists of a series of up-sampling operations to output an attention mask $\mathbf{\alpha}$ from $\mathbf{f}_{p}$: 
\begin{equation*}
	\alpha = \sigma(\mathbf{W}_{s}^{1}(\mathbf{D}(\mathbf{f}_{l}^{p})))
\end{equation*}
where $\mathbf{W}_{S}^{1} \in \mathbb{R}^{N\times N}$ are the parameters of a depth-wise separable 9$\times$9 convolution, the output attention map $\alpha$ implicitly contains the keypoints prior from action-specific knowledge.

On the other hand, inspired by the SENet~\cite{Hu_2018_CVPR}, the CR aims at performing channel-wise feature re-calibration in a global sense, where the per-channel summary statistics are utilized to selectively emphasis informative feature maps as well as suppress useless ones. To be specific, the latent feature passes through two fully-connected layers and a sigmoid activation to obtain an attention vector $\mathbf{\beta}$ for each channel
\begin{equation*}
	\beta = \sigma(\mathbf{W}_{c}^{2}\cdot\delta(\mathbf{W}_{c}^{1}\mathbf{f}_{l}^{p}))
\end{equation*}
where $\sigma(\cdot)$ and $\delta$ represent the sigmoid and ReLU functions respectively, $\mathbf{W}_{C}^{1}\in \mathbb{R}^{d \times N}$ and $\mathbf{W}_{C}^{1} \in \mathbb{R}^{N\times N}$ refer two fully-connection layers.

The channel-wise and spatial attention guidance is then applied to refine pose feature as 
\begin{equation*}
	\mathbf{\hat{F}}_{p} =\mathbf{F}_{p} \otimes(1 + \beta \otimes \alpha)
\end{equation*}

\subsubsection{Implementation Details}
\label{section:our-algorithm}

The HRNet~\cite{dhrn2019} pretrained on COCO is chosen as our backbone for pose feature extraction training. The proposed modules are appended after the last stage of HRNet. Our \textit{Encoder} and \textit{Decoder} use the corresponding downsample and upsample architecture in U-Net, respectively. For \textit{training}, the whole network is trained on the HiEve training set. For a fair comparison, same as we described in \ref{section:baseline-pose}, we take the Faster-RCNN~\cite{faster2015} as person detector. As the actions are annotated every 20 frames in HiEve, we utilize interpolation to create action category labels for all individuals in every frame. We set different learning rates for the backbone HRNet and our proposed modules, which are 1e-4 and 1e-3 respectively. In our experiments, we will show that our model gains the ability of mining potential action information to refine the poses. During training phase, the total loss for training is defined as:
\begin{equation*}
	\mathcal{L} = \mathcal{L}_{reg} + \mathcal{L}_{align}
\end{equation*}
where the $\mathcal{L}_{reg}$ is the traditional heatmap regression L2 loss. During inference, the action label embedding process is removed, and the proposed modules are connected with the last stage's output of HRNet. 

\begin{table*}[t]
	\centering
	\resizebox{\linewidth}{!}{
	\begin{tabular}{lccccccccccc}
		\hline
		Method     & \textbf{MOTA} & \textbf{w-MOTA} & \textbf{HOTA} & \textbf{MOTP} & \textbf{IDF1} & \textbf{MT} & \textbf{ML} & \textbf{FP} & \textbf{FN} & \textbf{IDSw} & \textbf{IDSw-DT}\\ \hline
		DeepSORT~\cite{deepsort2017}   &       27.12    &    21.95 &    25.25   &       70.47        &          28.55     &   8.50\%        &     41.45\%        &     5894        &   42668    &    2220  &    90 \\
		MOTDT~\cite{motda2018}      &       26.09     &  21.73  &    21.47 &          76.50     &    32.88           &          8.70\%   &      54.56\%       &      6318       &   43577       &   1599 &    76\\
		IOUtracker~\cite{ioutracker2017} &    \textbf{38.59}   &  \textbf{33.31}  &    \textbf{41.96}      &     76.23        &           38.62    &     28.33\%        &             27.60\% &    9640         &     28993  &  4153   &    92\\
		JDE~\cite{wang2019towards} &    33.12   &  27.78    &    30.63   &     72.27          &           36.01    &    15.11\%        &             24.13\% &    9526         &     33327  &  3747   &    93\\ 
		FairMOT~\cite{zhang2020fair} &    35.03   &  30.49    &    38.46   &    75.57          &           46.65    &     16.26\%        &             44.18\% &    6523         &     37750  &  995   &    79\\
		
		TPM~\cite{tpm2020} &    33.58   &  28.30    &    35.16    &    75.67        &         40.17 &	20.36\%	 & 29.80\% &	7395 &	31638 &	4536 &	94\\
		
		CenterTrack~\cite{centertrack2020} &    31.06   &  25.66    &    34.26   &    75.77         &         41.81 &	8.60\% &	27.91\% &	10014 &	35253 &	2767 &	94\\
		\hline
	\end{tabular}}
	\caption{Results of multi-person tracking baselines.}
\label{table:mot}
\end{table*}

\vspace{4pt}

\section{Experiments and results}
\label{section:main-results}
\subsection{Multi-person tracking}
\noindent
\textbf{Baselines}
\begin{itemize}
	\item DeepSORT~\cite{deepsort2017}. Based on the SORT~\cite{sort2016} algorithm, it extracts person appearance features by a pre-trained model, then simple nearest neighbor query is performed to track pedestrians.
	\item MOTDT~\cite{motda2018}. MOTDT  tackles unreliable detection by selecting candidates from outputs of both detection and tracks. Besides, a new scoring function for candidate selection is formulated by an efficient R-FCN.
	\item IOUtracker~\cite{ioutracker2017}.  IOUtracker proposes a very simple and efficient tracking algorithm, which only leverages the detection results and designs an IOU strategy to improve the performance of multi-objective tracking.
	\item JDE~\cite{wang2019towards}. JDE Tracker is the first joint pipeline for simultaneous detection and tracking, which produce the object embedding to accosiate persons across frames.
	\item FairMOT~\cite{zhang2020fair}. FairMOT is another joint detection-tracking pipeline, which focuses on addressing spatial misalignment with under an anchor-free manner.
	\item TPM~\cite{tpm2020}. TPM proposes a tracklet-plane matching process to model and reducing the interference from noisy or confusing object detections.
	\item CenterTrack~\cite{centertrack2020}. A simple but efficient method, which applies a detection model to a pair of images and detections from the prior frame.
\end{itemize}

\vspace{3pt}
\noindent
\textbf{Implementation Details}

\noindent
Faster R-CNN~\cite{faster2015} is used to obtain the public results of bounding-boxes firstly. In MOTDT and DeepSORT, we use the train set of HiEve and the ground truth to fine-tune the official deep models in these methods. Then, we evaluate them in the HiEve test dataset with the public detection results. The threshold of detections is set to be 0.2. 

\vspace{3pt}
\noindent
\textbf{Results and Analysis}

\begin{figure}[t]
	\centering
	\begin{minipage}[t]{0.9\linewidth}
		\centering
		\includegraphics[width=1\textwidth]{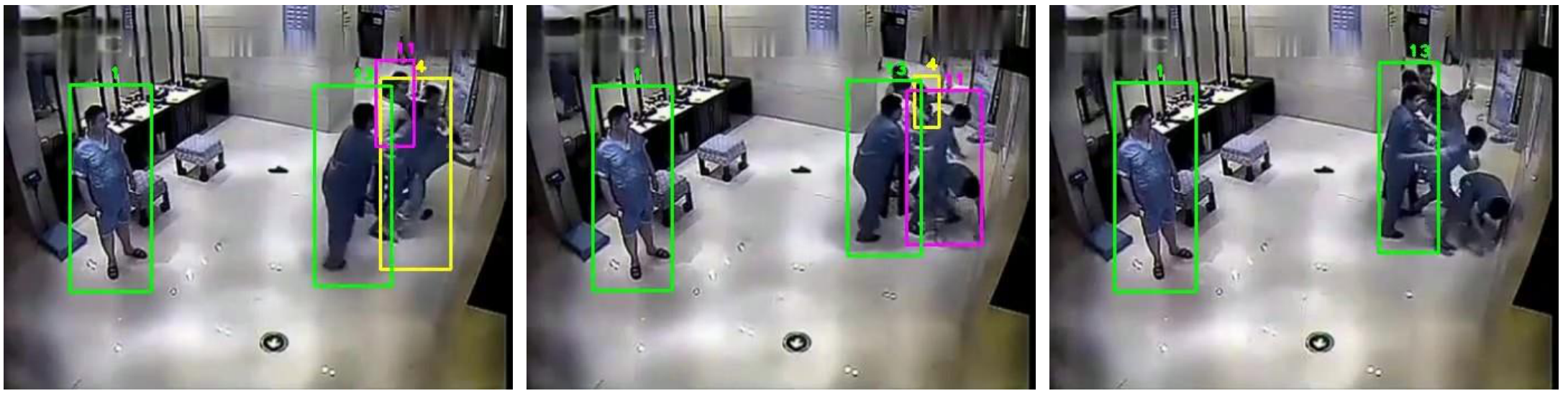}
		\subcaption{DeepSORT}
	\end{minipage}
	\begin{minipage}[t]{0.9\linewidth}
		\centering
		\includegraphics[width=1\textwidth]{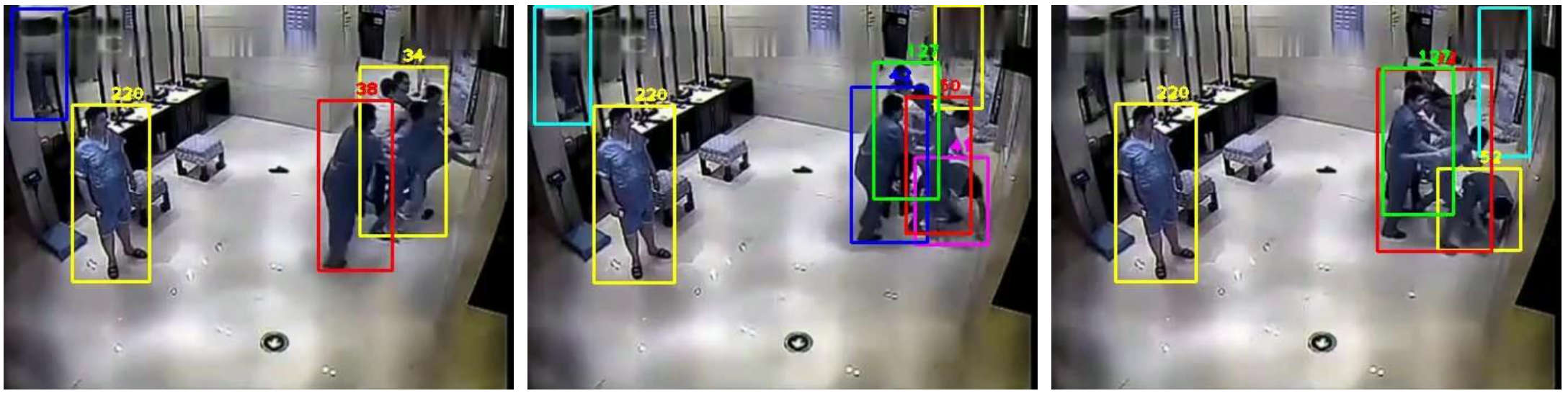}
		\subcaption{IOUtracker}
	\end{minipage}
	\begin{minipage}[t]{0.9\linewidth}
		\centering
		\includegraphics[width=1\textwidth]{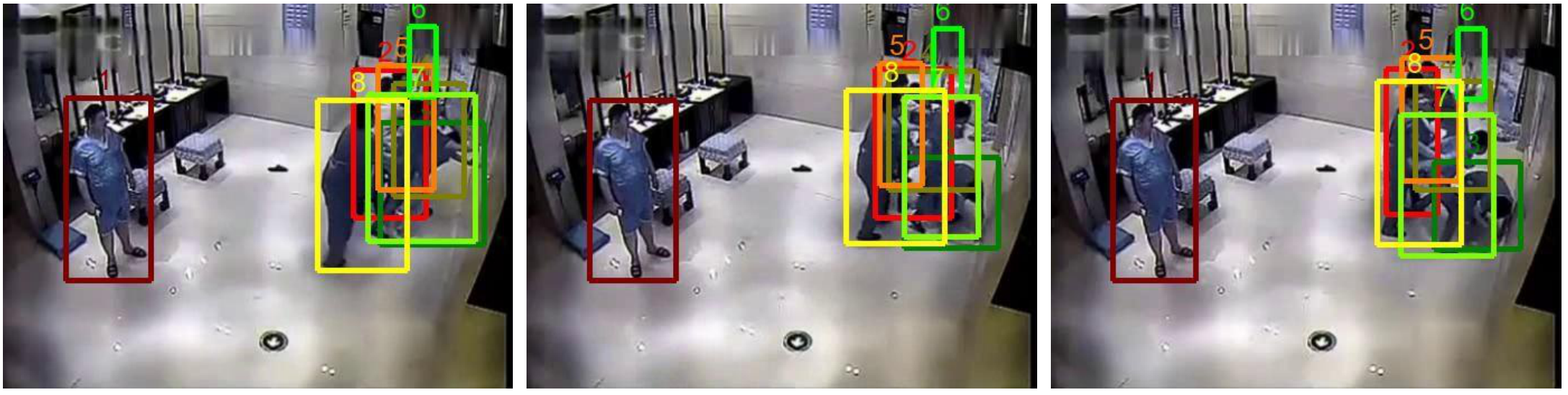}
		\subcaption{GT}
	\end{minipage}
	\caption{Visualized results of MOT baselines and the ground-truth (GT).}
	\label{fig:visualized_mot}
\end{figure}

\noindent
The results of these baselines are shown in \autoref{table:mot} and \autoref{fig:visualized_mot}. We can observe that all of their performances are not ideal. This is because our dataset has complex scenes and a large number of overlapping targets, making identification and tracking more difficult. IOUtracker~\cite{ioutracker2017} performs best on our dataset, while MOTDT~\cite{motda2018} and DeepSORT~\cite{deepsort2017} have relatively worse performance. Meanwhile, the joint detection-and-tracking solution JDE~\cite{wang2019towards}, CenterTrack, and FairMOT~\cite{zhang2020fair} also performs worse than the simple IOU Tracker. The reason is that HiEve contains numerous crowded scenes and occlusions, so it's hard to extract discriminative features to distinguish different object instances. 

\begin{table*}[ht]
	\centering
	\resizebox{\linewidth}{!}{
		\begin{tabular}{lcccccccc}
			\hline
			Method          & \multicolumn{1}{l}{\textbf{w-AP@avg}} & \multicolumn{1}{l}{\textbf{w-AP@0.5}} & \multicolumn{1}{l}{\textbf{w-AP@0.75}} & \multicolumn{1}{l}{\textbf{w-AP@0.9}} & \multicolumn{1}{l}{\textbf{AP@avg}} & \multicolumn{1}{l}{\textbf{AP@0.5}} & \multicolumn{1}{l}{\textbf{AP@0.75}} & \multicolumn{1}{l}{\textbf{AP@0.9}} \\ \hline
			DHRN~\cite{dhrn2019}            & 52.78                             & 61.73                                 & 50.73                                  & 45.91                                  & 56.40                            & 64.89                               & 54.56                                & 49.76                               \\
			Simple Baseline~\cite{simple2018}
			& 50.51                             & 59.90                                 & 47.90                                  & 43.74                                  &54.44                             & 63.56                              & 52.19                                & 47.59                        \\
			HigherHRNet~\cite{higherhrnet}
			& 22.03                             & 25.65                            & 21.37                            &19.06                             & 24.92                               & 28.74                                & 24.23
			&21.77\\
			
			RSN~\cite{RSN2020} 
			&52.25&	63.34	&49.75&	43.65	&55.46&	66.23&	53.24&	46.92	  	 \\
			
			DEKR~\cite{DEKR2021} 
			&47.46&	56.47	&44.87&	41.04&	49.42&	58.07&	47.09&	43.10  	 \\ 
			HRFormer~\cite{hrformer2021}
			&51.03 &	 60.77 &	 48.33 &	 44.00 &	 54.67	 & 64.07	& 52.21 &	 47.74	 \\ 
			
			Action-guided pose estimation (Ours)
			& \textbf{53.92}                             & 63.72                                 & 51.67                              & 46.36                                  & 57.68                            & 67.15                               & 55.60                                & 50.30                               \\ \hline
	\end{tabular}}
	\caption{Results of multi-person pose estimation.}
	\label{table:pose-e}
\end{table*}
\subsection{Multi-pose estimation}
\label{section:baseline-pose}
\noindent
\textbf{Baselines}
\begin{itemize}
	\item Simple-Baseline~\cite{simple2018}. It improves the performance of ResNet~\cite{resnet2016} backbone on pose estimation by adding a few deconvolutional layers. 
	\item DHRN~\cite{dhrn2019}. It aims to learn high-resolution representations for pose estimation. Specifically, the high-to-low resolution subnetworks are added one by one to form more stages. 
	\item HigherHRNet~\cite{higherhrnet} It's a bottom-up approach, which first detects all human keypoints with improved HRNet and then performs keypoints matching for each individual.
	\item DEKR~\cite{DEKR2021} It learns to directly regress different keypoints with distinctive adaptive convolutions, which could disentangle the representation for keypoints and obtain ideal performance under bottom-up paradigm.
	\item RSN~\cite{RSN2020} It devises a residual steps network to learn delicate local representations by  intra-level feature fusion.
	\item HRFormer~\cite{hrformer2021} It adopts the idea of multi-resolution parallel in DHRN~\cite{dhrn2019} to the Transformer~\cite{transformer2017} architecture.
	\item Ours. Our proposed action-guided pose estimation baseline.
\end{itemize}

\vspace{3pt}
\noindent
\textbf{Implementation Details}

\noindent
For the above top-down methods, we take the same detection results of Faster-RCNN~\cite{faster2015} as their input.  For all mentioned methods, we use their official codes to conduct implementation and experiments. Specifically, we download their public COCO pre-trained weights as initialization and further fine-tune them on our HiEve training set. We report their performance on our HiEve test set as the final results for a fair comparison.

\vspace{3pt}
\noindent
\textbf{Results and Analysis}

\begin{figure}
	\centering
	\begin{minipage}[t]{0.45\linewidth}
		\centering
		\includegraphics[width=1\textwidth]{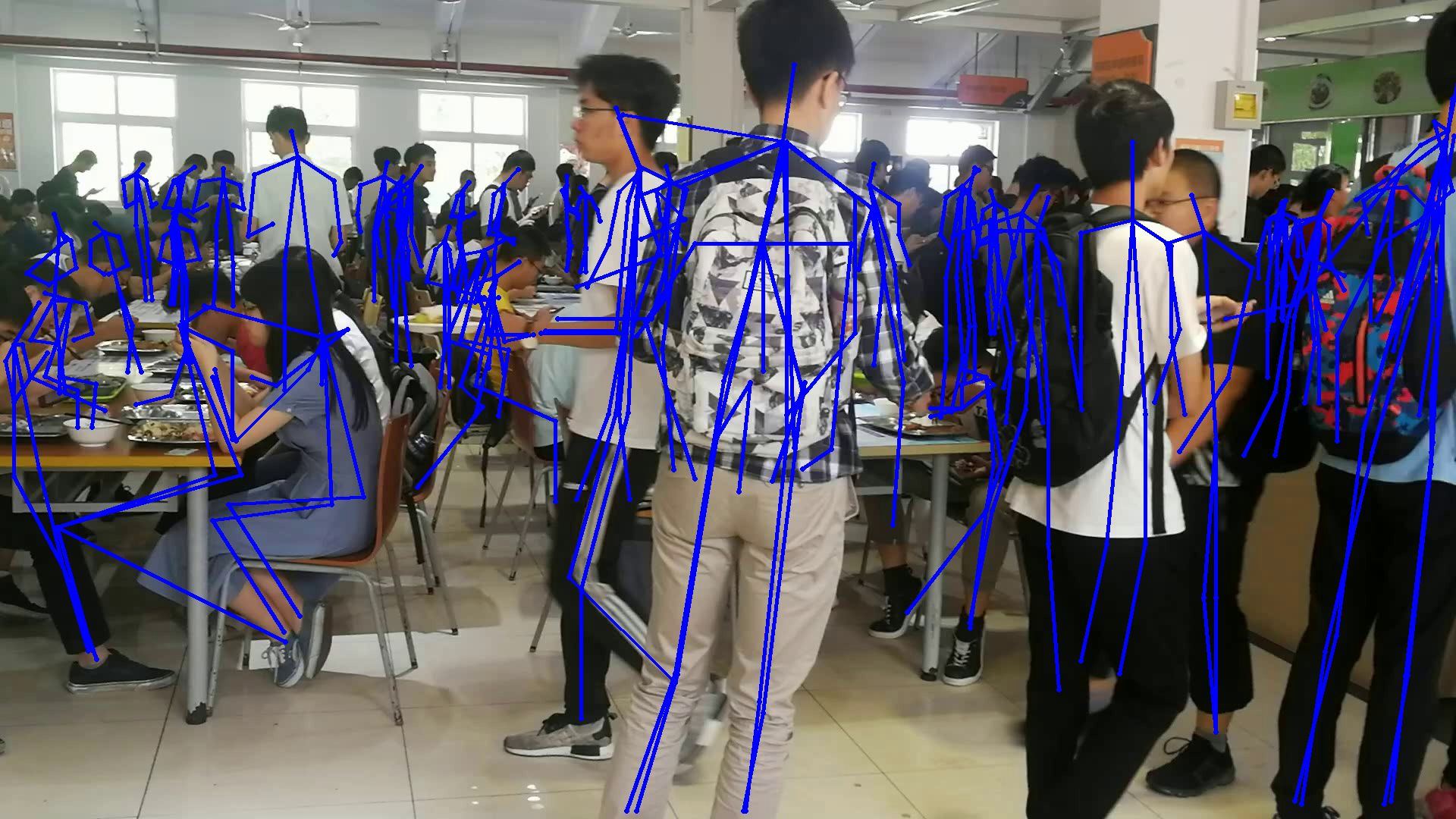}
		\subcaption{HRFormer}
	\end{minipage}
	\begin{minipage}[t]{0.45\linewidth}
		\centering
		\includegraphics[width=1\textwidth]{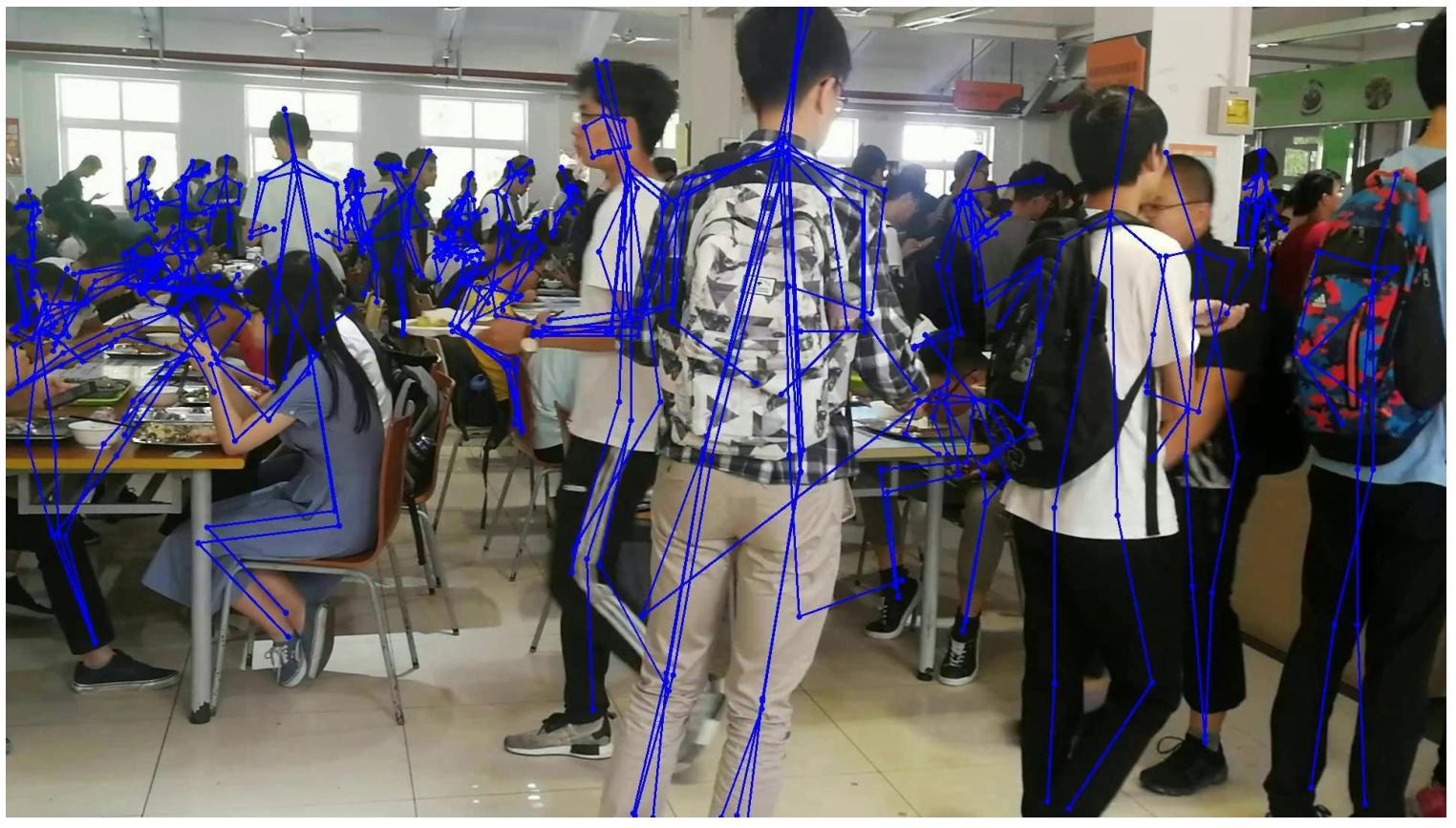}
		\subcaption{DHRN}
	\end{minipage}
	\begin{minipage}[t]{0.45\linewidth}
		\centering
		\includegraphics[width=1\textwidth]{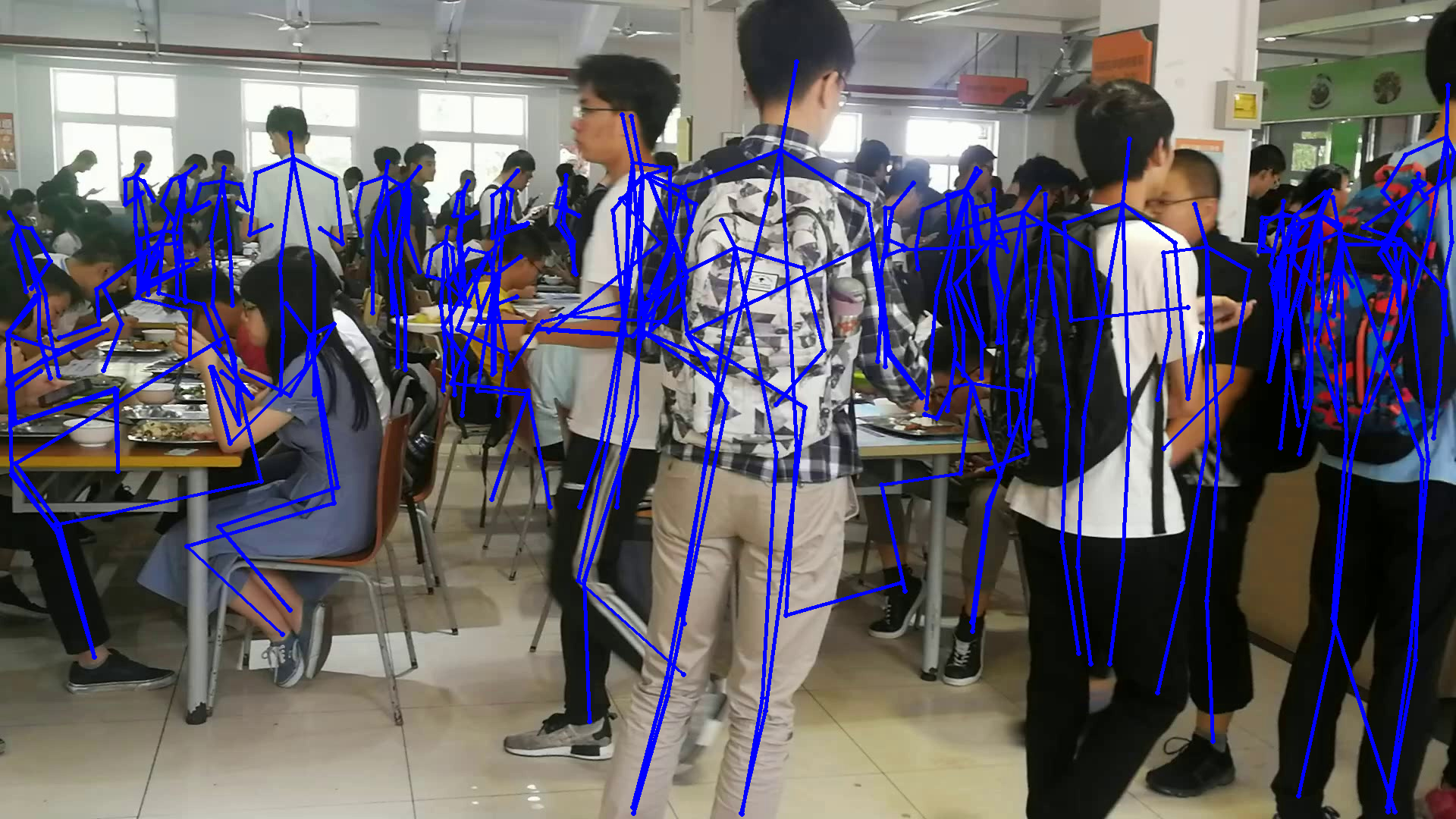}
		\subcaption{Ours}
	\end{minipage}
	\begin{minipage}[t]{0.45\linewidth}
		\centering
		\includegraphics[width=1\textwidth]{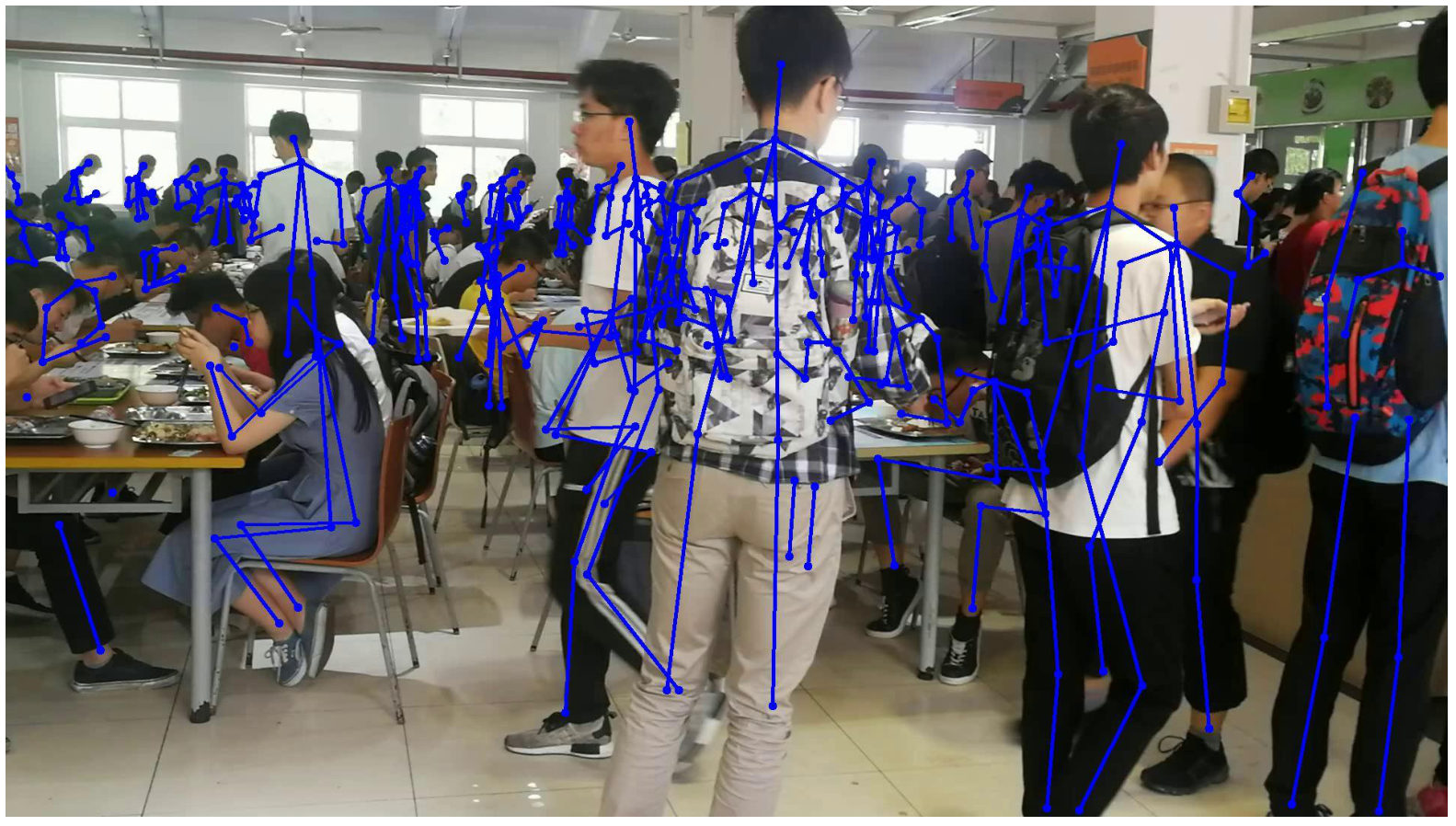}
		\subcaption{GT}
	\end{minipage}
	\caption{Visualized results of pose estimation baselines and the ground-truth (GT).}
	\label{fig:visualized_pose_estimation}
\end{figure}

\noindent
We present the evaluation results in \autoref{table:pose-e} and the visualization results in \autoref{fig:visualized_pose_estimation}.  It can be observed that DHRN~\cite{dhrn2019} performs best excluded our proposed method. Interestingly, the performance of recently proposed HRFormer~\cite{hrformer2021} falls between Simple-Baseline and DHRN. The reason is probably that transformer-based networks tend to overfit the training set. In fact, the performance of HRFormer on the validation set began to degrade earlier than other methods when we perform finetune on HiEve dataset. For bottom-up based methods, the recently proposed DEKR~\cite{DEKR2021} surpasses the HigherHRNet~\cite{higherhrnet} by a significant margin. The reason may be that the DEKR obtained disentangled representation for different keypoints using adaptive convolutions, which contributes to distinguishing the occlusion of human bodies. It can also be noticed that our proposed action-guided pose estimation further boosted the performance of DHRN by 1.13 w-AP. The comparisons manifest that by introducing action category information, our proposed simple baseline with aligned features and pose refine mechanisms could generate more accurate keypoint locations in crowded scenes. The success of this simple baseline also proves that leveraging the diverse annotation in the HiEve dataset could improve pose estimation.

\subsection{Pose tracking} 
\noindent
\textbf{Baselines}
\begin{itemize}
	\item PoseFlow~\cite{poseflow2018}. It's an efficient pose tracker based on flows and top-down approaches RMPE~\cite{alpha2017}. An online optimization framework is designed to build the association of cross-frame poses and form pose flows (PF-Builder). Then, a novel pose flow non-maximum suppression (PF-NMS) is designed to robustly reduce redundant pose flows and re-link temporal disjoint ones.

	\item LightTrack~\cite{lighttrack2019}. LightTrack is an effective light-weight framework for online human pose tracking. It unifies single-person pose tracking with multi-person identity association. 
	
	\item Our method + PoseFlow. Based on the pose estimation results of our algorithm, we adapted PoseFlow method to conduct human pose tracking across frames. 
\end{itemize}

\vspace{3pt}
\noindent
\textbf{Implementation Details}

\noindent
In LightTrack, \textit{YOLO v3}, Siamese GCN, and MobileNet are selected as the keyframe detector, ReID module, and pose estimator respectively. We use DeepMatching to extract dense correspondences between adjacent frames in PoseFlow. All weights of model inherit from pre-trained models on MSCOCO~\cite{coco}.

\begin{table}[t]
	\centering
	\begin{tabular}{lccc}
		\hline
		Method     & \textbf{MOTA} & \textbf{MOTP} & \textbf{AP} \\ \hline
		RMPE + PoseFlow~\cite{poseflow2018}   & 44.17        & 48.33         & 60.10       \\
		LightTrack~\cite{lighttrack2019} & 27.44         & 55.23         & 29.36       \\
		Ours + PoseFlow & \textbf{45.36}         & 49.97      & 63.16       \\\hline
	\end{tabular}
	\caption{Results of pose tracking baselines.}
	\label{table:pose-t}
\end{table}
\begin{figure}[t]
	\centering
	\begin{minipage}[t]{0.9\linewidth}
		\centering
		\includegraphics[width=1\textwidth]{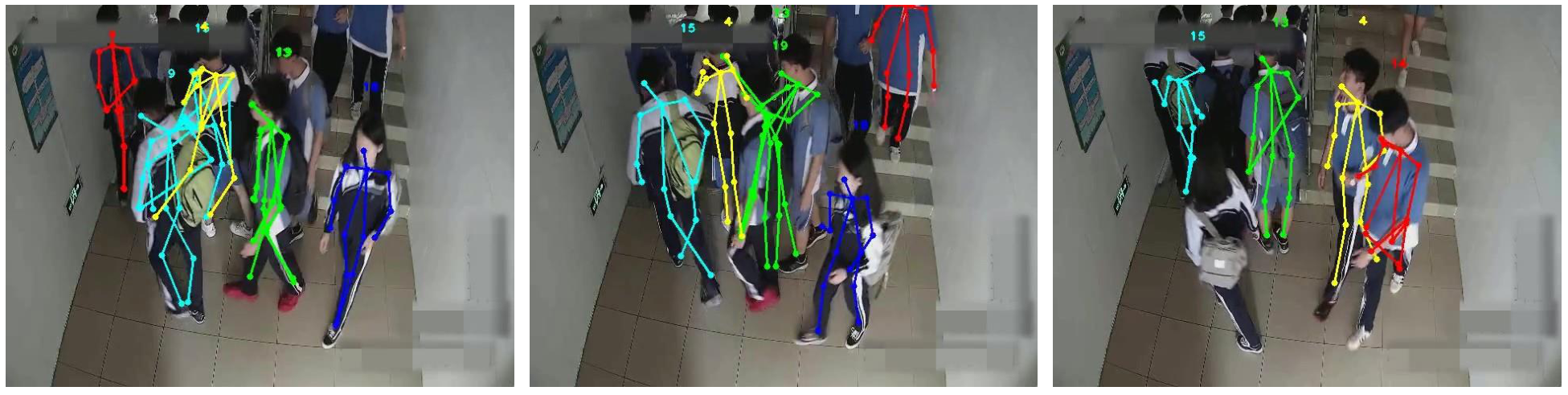}
		\subcaption{LightTrack}
	\end{minipage}
	\begin{minipage}[t]{0.9\linewidth}
		\centering
		\includegraphics[width=1\textwidth]{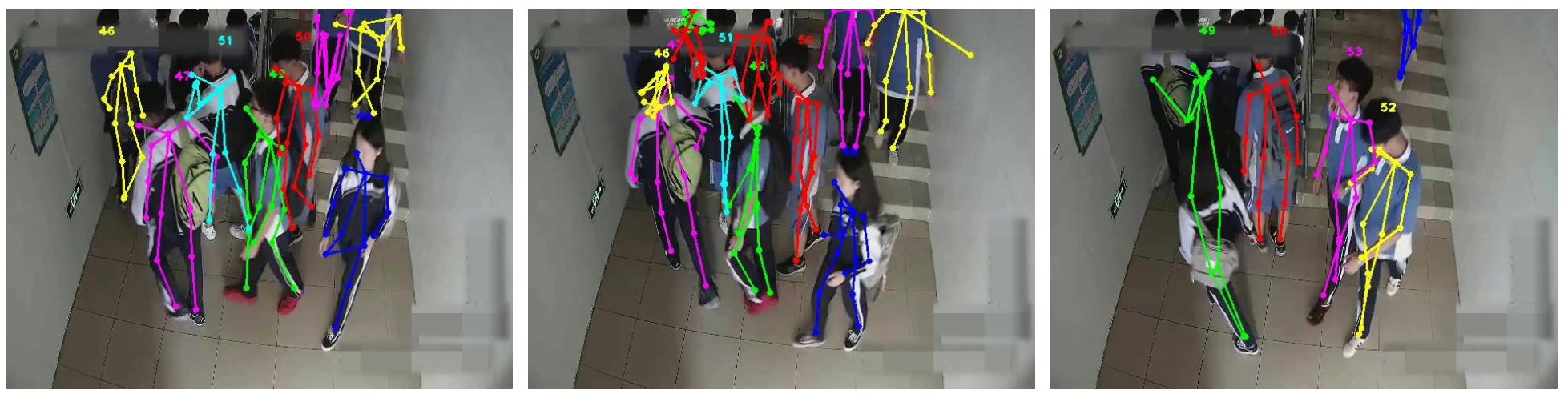}
		\subcaption{PoseFlow}
	\end{minipage}
	\begin{minipage}[t]{0.9\linewidth}
		\centering
		\includegraphics[width=1\textwidth]{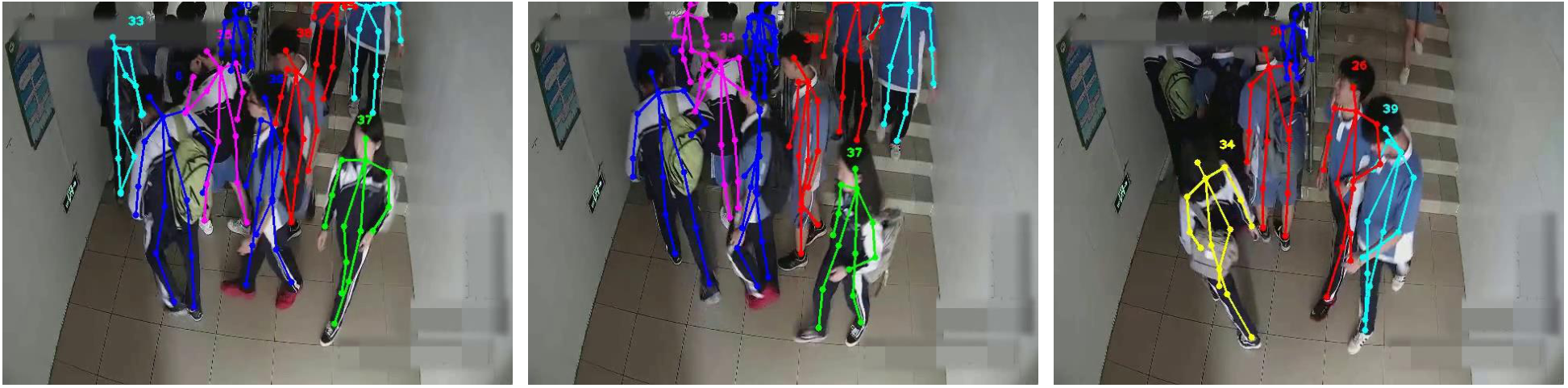}
		\subcaption{Ours + PoseFlow}
	\end{minipage}
	\begin{minipage}[t]{0.9\linewidth}
		\centering
		\includegraphics[width=1\textwidth]{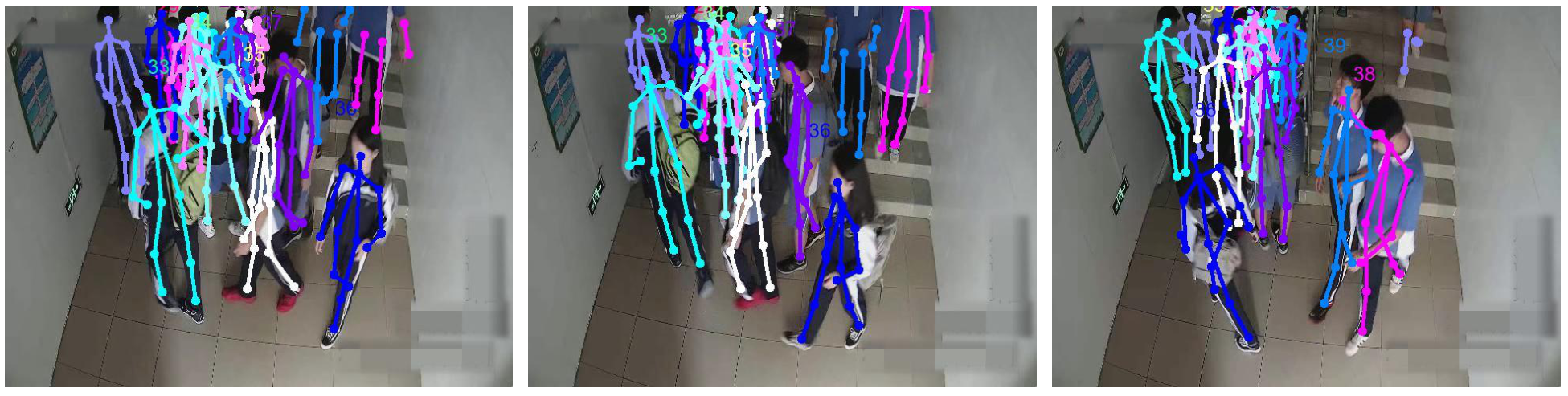}
		\subcaption{GT}
	\end{minipage}
	\caption{Visualized results of pose tracking baselines and the ground-truth (GT).}
	\label{fig:visualized_pose_tracking}
\end{figure}
\vspace{8pt}

\noindent
\textbf{Results and Analysis}

\noindent
The performance comparison of these three methods is presented in \autoref{table:pose-t}. As expected, the flow-based algorithm PoseFlow achieves higher performance while LightTrack~\cite{lighttrack2019} mainly aims to strike a balance between speed and accuracy. The \autoref{fig:visualized_pose_tracking} shows the visualization results of them, PoseFlow is able to track more people than LightTrack, but they all face the issue of losing objects and bad keypoints localization in crowded scenes. Enhanced by the accurate keypoints location of our proposed pose estimation algorithm, the performance of PoseFlow could be further improved.

\begin{table*}[ht]
	\centering
	\begin{tabular}{l|cccc|cccc}
		\hline
		Method& \multicolumn{4}{c|}{\textbf{wf-mAP}} & \multicolumn{4}{c}{\textbf{f-mAP}} \\\hline 
		Threshold           & \multicolumn{1}{c}{\textbf{avg}} & \multicolumn{1}{c}{\textbf{0.5}} & \multicolumn{1}{c}{\textbf{0.6}} & \multicolumn{1}{c|}{\textbf{0.75}} & \multicolumn{1}{c}{\textbf{avg}} & \multicolumn{1}{c}{\textbf{0.5}} & \multicolumn{1}{c}{\textbf{0.6}} & \multicolumn{1}{c}{\textbf{0.75}} \\ \hline
		I3D (RPN)~\cite{baseline_for_ava2018}         & 6.88                                & 9.65                                  & 7.91                                   & 3.07                                     & 8.31                              & 11.01                                 & 9.65                                  & 4.26                                   \\
		I3D~\cite{faster2015} & 10.13                               & 13.35                                  & 11.57                                  & 5.49                                     & 10.95                              & 14.50                                 & 12.33                                 & 6.01                                   \\
		VTN~\cite{videotransformer2019} & 7.28                                & 9.88                                   & 8.32                                   & 3.65                                     & 7.03                               & 9.32                                  & 8.10                                  & 3.66     \\  
		FeatureBank~\cite{featurebank2019} & 6.36                               & 8.69                                   & 7.21                                   & 3.19                                     & 8.42                               & 10.65                                  & 9.63                                  & 4.97     \\  
		
		LSTC~\cite{lstc2021}  &  7.44                              & 9.67                                      & 8.53                                  &  4.12                                    & 8.90                               &  11.36                               &10.54                                   &4.81  \\    
		SlowFast~\cite{slowfast}  & 12.08                               & 11.13                                      &                        12.84           &                                      12.27 &  14.12                              & 13.86                                &   14.75                                & 13.95 \\  
		Pose-aware action recognition (Ours)  & \textbf{13.16}                                &      12.35                                  & 13.56                                  & 13.58                                     & \textbf{14.90}                               & 14.10                                & 15.28                                  & 15.31                                   \\
		\hline  
		ST-GCN~\cite{stgcn} &  6.95                                  & 8.82                                  & 7.15                                     & 4.88                              & 7.69                                & 10.19                                 &  8.42 &      4.48                           \\
		\hline
		Video-Swin~\cite{video-swin}  & \textbf{15.67}                               & 18.62                                      &                        17.25           &                                      11.15 &  \textbf{18.78}                              & 20.26                                &   19.46                                & 16.61 \\  
		TimeSformer~\cite{timesformer}  & 14.18                               & 17.38                                      &                        14.22           &                                      10.94 &  17.43                              & 19.86                                &   17.75                                & 14.68 \\  
		\hline
	\end{tabular}
	\caption{Results of action recognition baselines.}
	\label{table:action}
\end{table*}
\subsection{Action recognition}
\noindent
\textbf{Baselines}
\begin{itemize}
	\item I3D (RPN)~\cite{baseline_for_ava2018}. In this method, the I3D~\cite{kinetics2017} network is applied for feature extraction and classification, and the feature from the labelled key-frame is fed to RPN~\cite{faster2015} for region proposal.
	\item I3D~\cite{baseline_for_ava2018}. We further improve the baseline in~\cite{baseline_for_ava2018} for better localization. To be specific, the Faster R-CNN detector~\cite{faster2015} is applied on the input key-frame to obtain the bounding box proposals. 
	\item VTN~\cite{videotransformer2019}. The VTN (Video Transformer Network) takes the I3D network as backbone and applies a key-value attention mechanism to model the interaction among objects before the classification layer to improve recognition results.
	\item FeatureBank~\cite{featurebank2019} It builds a long-term feature bank to store and update temporal features across frames to provide a global perception of videos.
	\item LSTC~\cite{lstc2021} It addresses the atomic action detection issue by modeling the action temporal reliance from shot-term and long-term context.
	\item SlowFast~\cite{slowfast}. The SlowFast model involves two pathway, the slow pathway operates at low frame rate, to capture spatial semantics, and the fast pathway operates at high frame rate, to capture motion at fine temporal resolution. 
	\item Ours. Our proposed pose-aware action recognition baseline.
	\item ST-GCN~\cite{stgcn2018} A skeleton-based action recognition method, leveraging GNNs to model the complex spatial-temporal relationships among human joints.
	\item TimeSformer~\cite{timesformer} The TimeSFormer is an  Transformer-based model, specifically developed for video understanding tasks, which excels in spatial-temporal modeling and action recognition across a diverse range of datasets.
	\item Video-Swin~\cite{video-swin}  It is a state-of-the-art Transformer-based approach specifically designed for video analysis tasks, showcasing remarkable performance across a wide range of video benchmarks.
\end{itemize}

\vspace{3pt}
\noindent
\textbf{Implementation Details}

\noindent
For all baselines except for SlowFast~\cite{slowfast}, we adopt the RGB-I3D~\cite{kinetics2017} network with Inception-V1, initialized with Kinetics-pretrained weights, as a video feature extractor. The SlowFast takes pretrained inflated-ResNet50~\cite{Wang_2018_CVPR} as backbone. In RPN+I3D, following~\cite{baseline_for_ava2018}, we generate region proposals by RPN on key-frame feature and implement action classification and box regression with I3D head. In Faster R-CNN+I3D and SlowFast, we use detection results of a Faster R-CNN detector as ROIs and perform action classification on RoI aligned features. In VTN, we use the same Faster R-CNN detection results as RoIs, but employ the transformer head in~\cite{videotransformer2019} for action classification. For ST-GCN, follow the \cite{stgcn}, we utilize its official toolbox\footnote{\url{https://github.com/yysijie/st-gcn}} to generate skeleton locations for frames using OpenPose. For Video-Swin, we select the \textit{Swin-B}\footnote{\url{https://github.com/SwinTransformer/Video-Swin-Transformer}} model pretrained on Kinetics-400 as the classification model. In terms of the TimeSformer, we adopt the standard TimeSformer model\footnote{\url{https://github.com/facebookresearch/TimeSformer}} pretrained on Kinetics-400 as the classification backbone.

\vspace{3pt}
\noindent
\textbf{Results and Analysis}

\begin{figure}[t]
	\centering
	\begin{minipage}[t]{0.9\linewidth}
		\centering
		\includegraphics[width=1\textwidth]{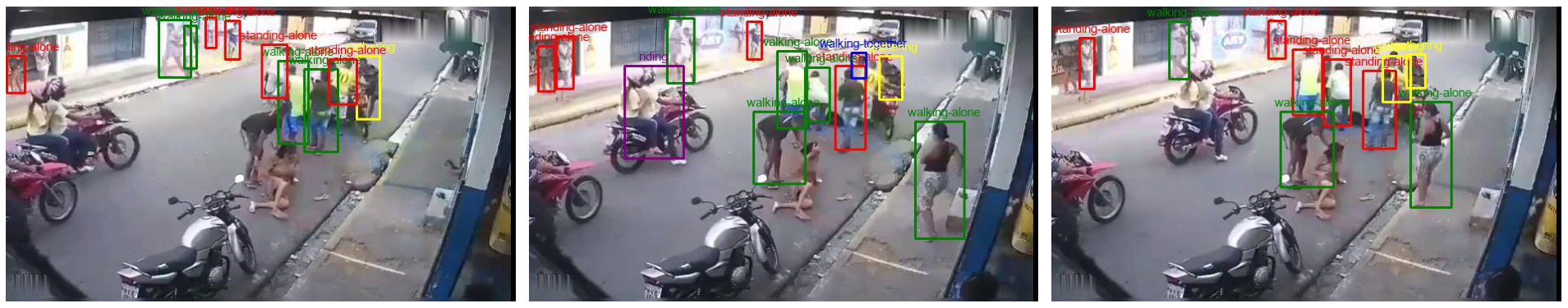}
		\subcaption{I3D (RPN)}
	\end{minipage}
	\begin{minipage}[t]{0.9\linewidth}
		\centering
		\includegraphics[width=1\textwidth]{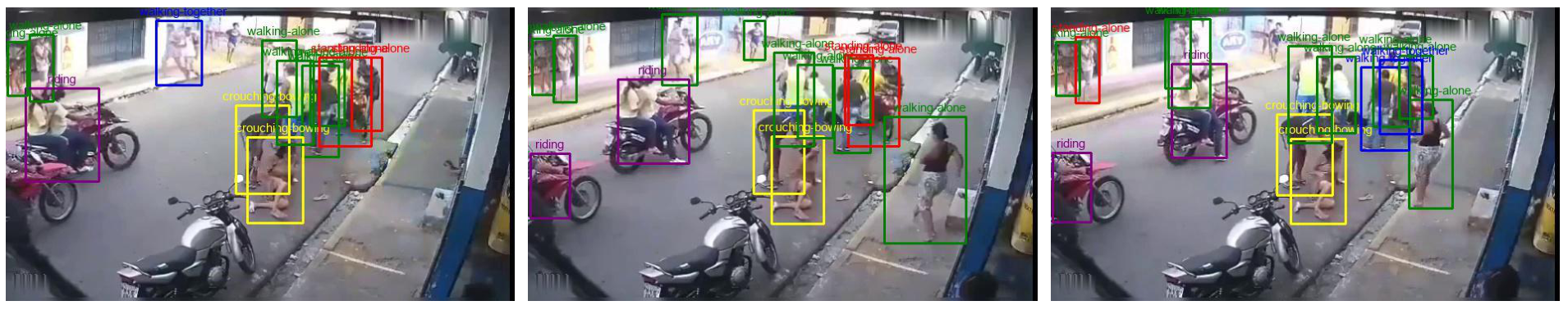}
		\subcaption{I3D}
	\end{minipage}
	\begin{minipage}[t]{0.9\linewidth}
		\centering
		\includegraphics[width=1\textwidth]{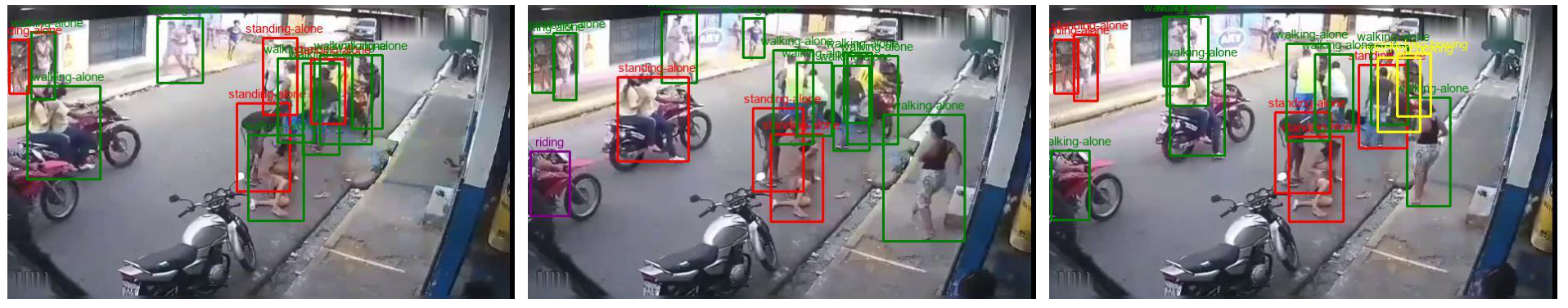}
		\subcaption{VTN}
	\end{minipage}
	\begin{minipage}[t]{0.9\linewidth}
		\centering
		\includegraphics[width=1\textwidth]{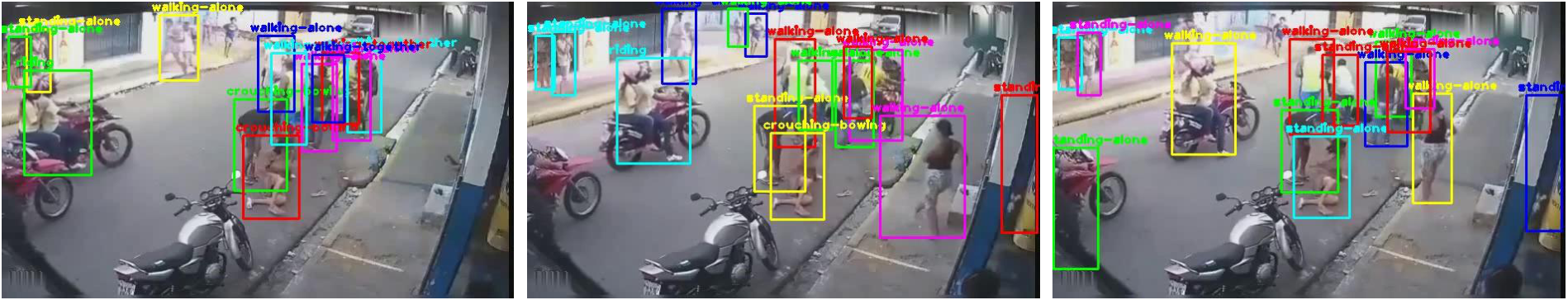}
		\subcaption{SlowFast}
	\end{minipage}
	\begin{minipage}[t]{0.9\linewidth}
		\centering
		\includegraphics[width=1\textwidth]{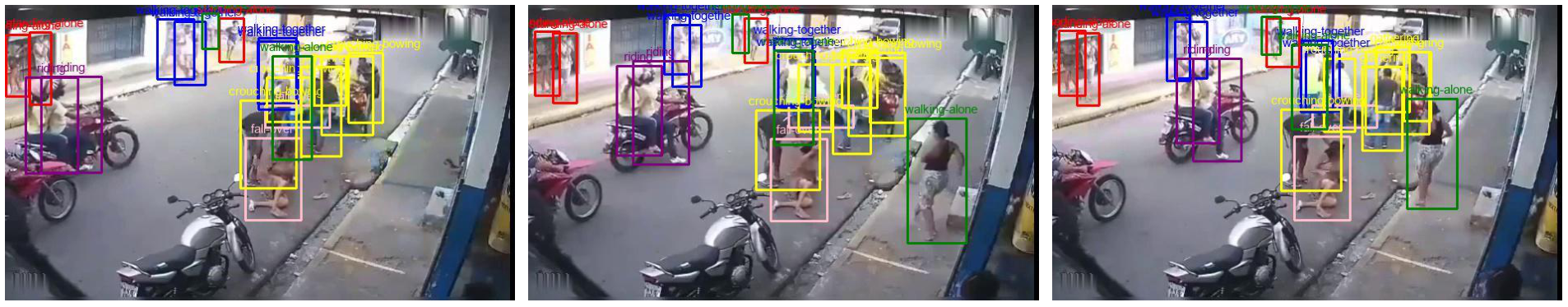}
		\subcaption{GT}
	\end{minipage}
	\caption{Visualized results of action recognition baselines and the ground-truth (GT).}
	\label{fig:visualized_action}
\end{figure}

\noindent
The main results are shown in \autoref{table:action}. The model employing I3D~\cite{kinetics2017} with Faster R-CNN detector performs best on our dataset, outperforming that using I3D for both detection and classification. It's probably because our dataset contains many crowded scenes, which is challenging for the detection stream. Therefore, utilizing a high-quality detector could significantly improve the detection performance. 
VTN~\cite{transformer2017} is superior on AVA~\cite{ava2018} dataset but performs comparatively poor on our dataset. Meanwhile, both the FeatureBank and LSTC can also perform great on AVA by virtue of their feature memory mechanism. However, their performance in HiEve is not satisfying as the AVA dataset. The reason might be that the AVA dataset focuses on human-human and human-object interaction, while our dataset pays more attention to the individual action under complex event conditions. Moreover, the visualization results of first three baselines are shown in \autoref{fig:visualized_action}, we can observe that it's difficult for these popular methods to recognize the anomalous actions in our dataset and none of them can tackle the prediction in crowded scenes well. The SlowFast owns the best performance in HiEve excluded the Transformer-based methods. Nevertheless, our proposed simple action recognition baseline still surpasses the vanilla Slowfast with 1.08 wf-mAP and 0.78 f-mAP. The difference in improvement on these two metrics indicates that combining the pose motion pattern can better address the action recognition under crowded scenes. The success of this simple baseline also proves that leveraging the diverse annotation in the HiEve dataset could improve the action recognition task.  In terms of the Transformer-based methods (Video-Swin and TimeSformer), they significantly outperform all the above baselines, which is consistent with their great performance on other action detection datasets. Specifically, the performance of the Video-Swin model surpasses our proposed baseline (based on the SlowFast model) and achieves the best results. These findings demonstrate that more powerful long-term spatial-temporal modeling is beneficial for action recognition in our HiEve dataset. As for the skeleton-based method ST-GCN, we can observe that it is not ideal compared to most RGB-based methods. This can be attributed to the difficulty of obtaining accurate pose estimations in our HiEve dataset due to heavy occlusion and complex scenes. In contrast, commonly-used skeleton-based action recognition datasets (e.g., NTU-RGB+D dataset~\cite{ntu}) feature fixed and simple scenes (indoor settings with only a single person), allowing for relatively accurate pose estimation for subsequent action recognition. Furthermore, these observations also validate the rationality of our proposed method, which leverages ground-truth skeletons as auxiliary information during training to enhance the RGB-based action recognition backbone. This paradigm enables us to utilize pose information for action recognition while simultaneously avoiding inaccurate pose estimation.

\section{More Analysis and ablation study}
\begin{figure*}[ht]
	\centering
	\hspace{-5mm}
	\begin{minipage}[t]{0.4\linewidth}
		\includegraphics[width=1\linewidth]{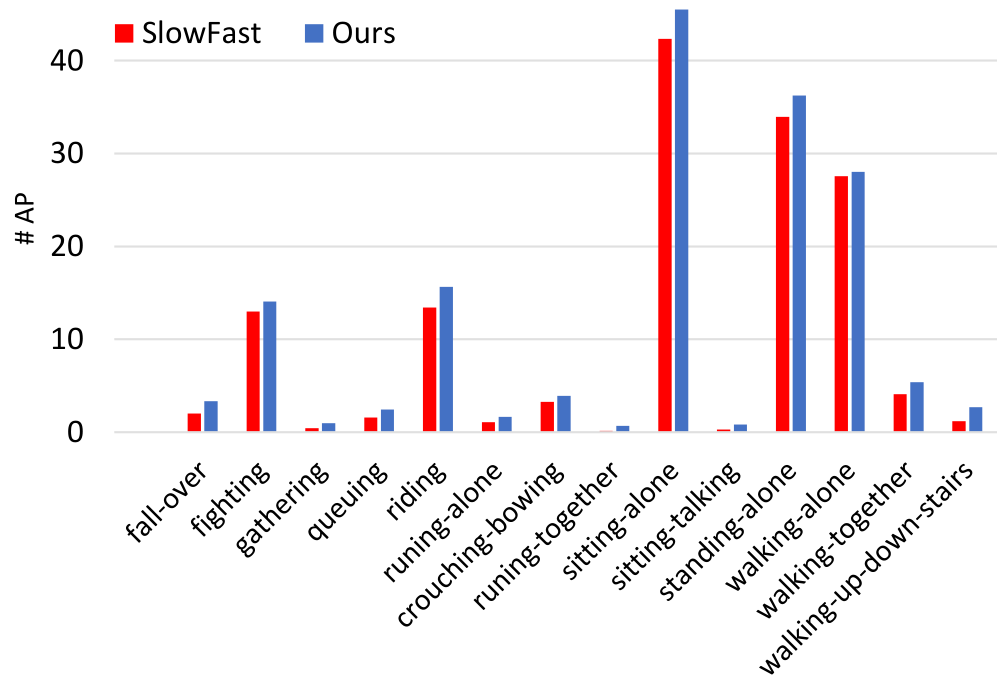}
		\caption{The performance of SlowFast and ours on each action category in HiEve}
		\label{fig:slowfast_category}
	\end{minipage}
	\hspace{0.5cm}
	\begin{minipage}[t]{0.4\linewidth}
		\includegraphics[width=1\linewidth]{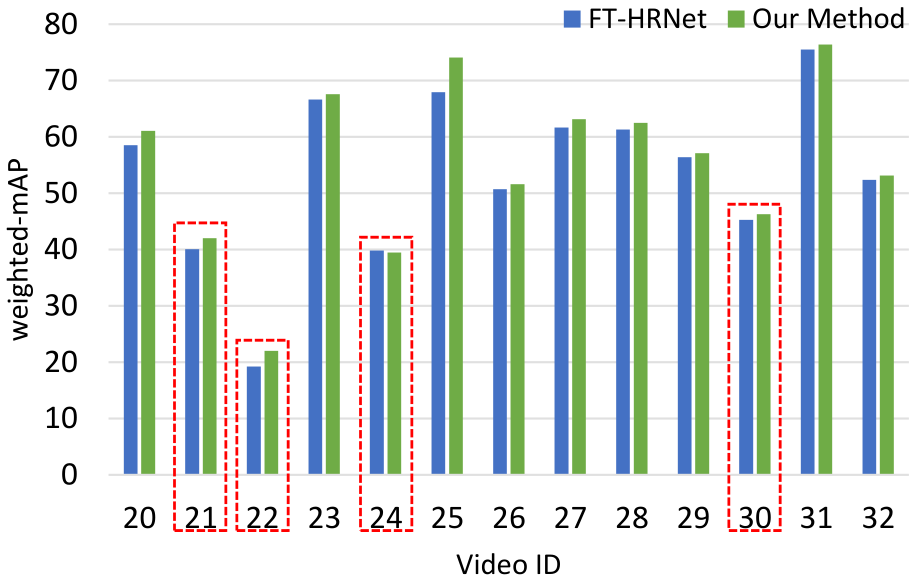} 
		\caption{The performance of FT-HRNet on each video sequence in HiEve. Hard video examples (weighted-AP $\leq$ 50) are emphasized by red dashed boxes.}
		\label{fig:poes_resutls_each_video}
	\end{minipage}
\end{figure*}
In this section, we first conduct experiments to analyze the characteristics of our HiEve dataset. Then, the ablation studies of our proposed algorithm will be presented to evaluate different variants of our proposed algorithm.
\subsection{Experimental characteristics}
\begin{table*}
	\centering
	\begin{minipage}[t]{0.48\linewidth}
		\centering
		\begin{tabular}{cc|cc}
			\hline
			\multicolumn{2}{c|}{Modules} & \multicolumn{2}{c}{Performance} \\ \hline
			ADAM          & PRM          & w-AP@avg          & AP@avg            \\ \hline
			&              & 52.78         & 56.40          \\
			\checkmark             &              & 53.10          & 56.87          \\
			\checkmark            &       \checkmark       & \textbf{53.92}         & \textbf{57.68}         \\ \hline
		\end{tabular}
			\caption{Results of breakdown modules of our algorithm on HiEve dataset. \checkmark means the module is used}
			\label{table:breakdown}
	\end{minipage}
	\hspace{0.3cm}
	\begin{minipage}[t]{0.48\linewidth}
		\centering
		\begin{tabular}{cc|cc}
			\hline
			\multicolumn{2}{c|}{Refinement Setting} & \multicolumn{2}{c}{Performance}  \\ \hline
			\multicolumn{1}{c|}{SR}       & CR      & \multicolumn{1}{c|}{w-AP@avg} & AP@avg \\ \hline
			\checkmark &         & 53.20                           & 56.97    \\
			&   \checkmark      & 53.65                           & 57.25    \\
			\checkmark     &   \checkmark      & \textbf{53.92}                           & \textbf{57.68}    \\ \hline
		\end{tabular}    
	\caption{Results by different refinement configurations}
	\label{table:attention}
	\end{minipage}
\end{table*}

\begin{table}[h]
	\centering
	\resizebox{\linewidth}{!}{
		\begin{tabular}{c|c|c}\hline
			\multirow{2}{*}{Pretraining ?}   & \multicolumn{2}{c}{Downstream task} \\\cline{2-3}
			& HRNet~\cite{dhrn2019} on COCO & MODT~\cite{motda2018} on MOT20 \\\hline
			NO & 74.4 & 46.4 \\
			YES & \textbf{74.8} & \textbf{47.6} \\\hline
	\end{tabular}}
	\caption{Downstream task results with and without HiEve pretraining}
	\label{table:transfer}
\end{table}

\textbf{Group \& fine-grained action} First, to better understand the difficulty of action recognition on the HiEve, we calculate the per-class AP value for each action category. \autoref{fig:slowfast_category} displays the results obtained by SlowFast~\cite{slowfast}. What stands out in this figure is the poor performance of some group behavior recognition, such as \textit{`gathering'}, \textit{`running-together'}, and \textit{`sitting-talking'}. Besides, the performance encounters a marked decline when recognizing fine-grained actions. For example, it's hard to distinguish the \textit{`running-alone'} from \textit{`walking-alone'}.  Compared to the vanilla SlowFast, our proposed action recognition baseline can effectively improve the accuracy of categories highly related to human skeletons. We also notice that our proposed baseline only gains slight improvement in these group-level and fine-grained categories. These results suggest that introducing pose information does improve action recognition under complex scenes. However, in our future work, specific measures need to be taken to further boost the performance of fine-grained \& group action categories in the HiEve dataset.

\noindent\textbf{Hard video sequence} First, we make a simple subjective analysis of the test video sequence. The \textit{CrowdIndex} is calculated for each test video sequence to measure the crowding level of frames. The top-3 sequences with the highest \textit{CrowdIndex} could be naturally regarded as relatively hard examples in the test set. Specifically, they are \textit{hm\_in\_bus} (ID:21), \textit{hm\_in\_dining\_room2} (ID:22), and \textit{hm\_in\_subway\_station} (ID:24). Furthermore, we report the weighted-AP of FT-HRNet\cite{dhrn2019} on each video sequence, since this metric pays more attention to crowded scenarios. As shown in \autoref{fig:poes_resutls_each_video}, consistent with our assumption, the performance shows a sharp degradation in all of these three video sequences. This indicates that the crowded level is a major influence on video understanding tasks in HiEve. Surprisingly, the performance on video sequence \textit{hm\_in\_stair3} (ID:30) also meets a marked drop whereas its crowded level is relatively low among all sequences. The reason for this is that it was dominated by the overhead view. To sum up, the hard example in our data set are close to the real-world scenes, namely, the severe human occlusion and various video angles.

\noindent\textbf{Upper bound test}
All the human-centric video understanding tasks are tightly associated with object detection. 
To study the impact of detection accuracy in the HiEve dataset, we conduct the upper bound test on each task with specific oracle models, where the ground-truth bounding-boxes are directly used during testing, including multi-person tracking, pose estimation, and action recognition. We compared them with the normal setting that we described in \autoref{section:main-results} without ground-truth. \autoref{table:upper_bound} lists the upper bound results for each track.
It suggest that the tasks requiring temporal reasoning (Track1\&3\&4) rely more on the accuracy of the detection. In contrast, the pose estimation track is more dependent on the corresponding algorithm than the detection results.

\noindent\textbf{Ability for knowledge transfer} HiEve covers large amounts of video frame data with a wide range of human-centric annotations, making it well suitable for model pretraining to inject these models with more comprehensive prior knowledge on downstream tasks. To demonstrate it, we conduct experiments on transfer learning from HiEve to other two related downstream tasks, human pose estimation and multiple object tracking. In detail, we apply HRNet~\cite{dhrn2019} for pose estimation on COCO~\cite{coco} and MOTDT~\cite{motda2018} on MOT20~\cite{mot2020}. For each task, we compare the results with and without pretraining on our HiEve datasets in \autoref{table:transfer}. For COCO we report the average AP value, for MOT20 we report the MOTA metric. It can be seen that for both downstream tasks, pretraining on HiEve can help improve the methods obtain better performance. 

Nevertheless, we can further observe a notable disparity in improvements between the two datasets, with a marginal improvement (0.4 AP) in COCO and a significant (1.2 MOTA) improvement in MOT20.  Our HiEve primarily offers prior knowledge for recognition in complex scenes compared to existing datasets.  Therefore, the contribution of pretraining on HiEve is related to the complexity of the downstream datasets.  Since the COCO dataset predominantly consists of simple and uncrowded scenes, it is reasonable that knowledge transferred from HiEve to COCO yields modest improvements. Conversely, the MOT20 dataset includes more challenging and crowded scenes compared to COCO, so we can see more significant improvement.

\begin{table}[t]
	\centering
	\begin{tabular}{cc|cc}
		\hline
		Track              & Methods                     & Normal     & Oracle               \\ \hline
		\multirow{2}{*}{1-human tracking} & \multirow{2}{*}{IOUTracker\cite{ioutracker2017}} & \multicolumn{2}{c}{MOTA}         \\ \cline{3-4} 
		&                             & 38.59      & \textbf{97.70}      \\ \hline
		\multirow{2}{*}{2-pose estimation} & \multirow{2}{*}{DHRN\cite{dhrn2019}}   & \multicolumn{2}{c}{w-AP@avg} \\ \cline{3-4} 
		&                             & 52.78      & \textbf{53.34}     \\ \hline
		\multirow{2}{*}{3-pose tracking} & \multirow{2}{*}{PoseFlow\cite{poseflow2018}}   & \multicolumn{2}{c}{MOTA}         \\ \cline{3-4} 
		&                             & 44.17      & \textbf{73.84}      \\ \hline
		\multirow{2}{*}{4-action recognition} & \multirow{2}{*}{SlowFast\cite{slowfast}}   & \multicolumn{2}{c}{wf-mAP@avg}    \\ \cline{3-4} 
		&                             & 12.08           &    \textbf{13.21}                 \\ \hline
	\end{tabular}
	\caption{The upper bound and normal setting results}
	\label{table:upper_bound}
\end{table}

\begin{table}[t]
	\centering
	\begin{tabular}{ccc|c}
		\hline
		\multicolumn{3}{c|}{Combination} & \multirow{2}{*}{wf-mAP@avg} \\ \cline{1-3}
		$m_1$       & $m_2$      & $m_3$      &                              \\ \hline
		\checkmark        &         &         &   12.36                           \\
		& \checkmark       &         &  12.58                            \\
		&         & \checkmark       &  12.44                            \\
		\checkmark        & \checkmark       &         &         12.79                     \\
		& \checkmark       & \checkmark       &     12.61                         \\
		\checkmark        &         & \checkmark       &         12.57                     \\
		\checkmark        & \checkmark       & \checkmark       &     \textbf{13.16}                         \\ \hline
	\end{tabular}
	\caption{Using features from different levels to predict. $m_l$ denotes feature output by stage-$l$ in ResNet-50.}
	\label{table:multi-level}
\end{table}

\subsection{Ablation study on our proposed baselines}
\subsubsection{Study on pose-aware action recognition}
The multi-level feature prediction task enables the video network to learn the pose-specific motion patterns in the training and testing phase. In this section, we aim to reveal the influence of multi-level feature selection. As shown in \autoref{table:multi-level}, we test different combinations of features across model stages to predict the pose-aware motion pattern. We can observe that using a single-level video feature is hard to conduct a precise prediction and only lead to a slight improvement. We also notice that the middle-level feature $m_2$ is crucial in multi-level feature joint prediction. The reason may be that the middle-level feature contains both high-level semantic information and low-level texture, which is beneficial for learning the pose-aware patterns. The performance reaches its peak when we combine all the features from three stages to conduct prediction. 

\subsubsection{Study on action-guided pose estimation}
The contributions of different modules in our model are first analyzed via experiments. \autoref{table:breakdown} presents the breakdown results of the action-guided domain alignment (ADAM) and pose refinement module (PRM). We can observe that by introducing action category information as a kind of regularization, the performance can achieve a large improvement of 1.24 weighted-AP. Besides, the performance can be further boosted to 54.00 w-AP with the refinement module, which indicates that the attention mask generated by the aligned latent feature fosters the pose feature revision and refinement.

\begin{figure}[t]
	\centering
	\includegraphics[width=0.95\linewidth]{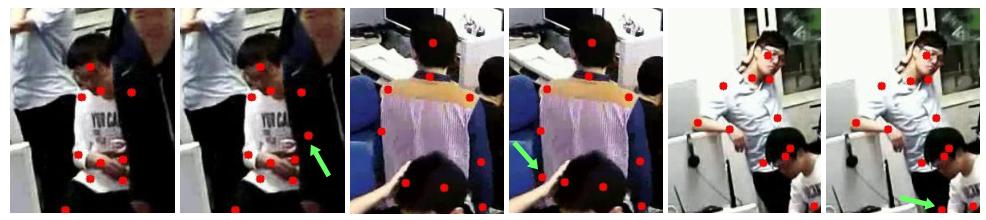}
	\caption{Prediction of keypoints in a test video without (\textit{left})/with (\textit{right}) PRM. Keypoints rectified by PRM are indicated by \textit{green} arrow. }\label{fig:pose_refine}
\end{figure}

To further validate the effectiveness of the PRM, we first visualize the pose estimation results without/with PRM module. As presented in \autoref{fig:pose_refine}, PRM is able to rectify the position of some keypoints or replenish some hard keypoints that are not detected.
Moreover, we also apply the SR and CR separately. As shown in Figure \autoref{table:attention}, each refinement plays an important role in the final performance. The application of single SR module gains 1.32 w-AP and 1.29 AP from the vanilla HRNet. 
With the combination of CR, the refinement module could provide the best performance. The contribution comparison demonstrates that the channel-wise refinement contributes more significantly to pose estimation refinement in crowded scenarios, which may be due to the difficulty of spatial attention modeling for severe occlusion scenes.

\subsection{Analysis of our proposed metrics}
\subsubsection{Will they leak any information about GT?} Note that the detailed weights and parameters for our three weighted metrics are not available to the researcher. All evaluations are conducted on the HiEve online server. The only way researchers can do for improving performance on weighted metrics is by exploring efficient methods or modules to handle complex events (such as crowded scenes, and anomaly action) in our video. 

\subsubsection{How do they contribute to a comprehensive comparasion?}
Our proposed weighted metrics aim to provide a comprehensive evaluation for various algorithms, especially their performance in real-world complex events. In most cases, the rank under these three metrics is consistent with the traditional metrics (as shown in \autoref{table:mot}, \autoref{table:pose-e}). However, when methods reach high performance with traditional metrics in HiEve, their performances will be too close to provide a fair comparison between them. Under this kind of condition, our proposed metrics could provide a comprehensive evaluation and comparison among these SOTA methods or submissions. And we'll show some real examples to further validate this.

\autoref{table:submission} presents submissions that selected from our public leaderboard on the HiEve website. As for the tracking task, we can observe that the submission `\textit{JiaRen.AI}' have a very close AP with submission `\textit{Commander}'. However, the  `\textit{JiaRen.AI}' marginally surpasses the `\textit{Commander}' on the w-MOTA. Our w-MOTA pays more attention to performance on disconnected tracks, which is a common problem in complex real-world scenes. Therefore, our leaderboard could provide a fair rank for these two methods and proves that the `\textit{JiaRen.AI}' is a better choice for MOT task in complex scenes. Our proposed metric `wf-mAP', which focuses more on frames with crowded or complex scenes, also contribute to a fair comparasion among action recognition methods. It can be seen from \autoref{table:submission} that the submission `\textit{CF}' outperforms the submission `\textit{8A}' with a significant margin in the traditional frame-mAP metric. However, these two methods have similar performance on our wf-mAP metric. It demonstrates that the performance of `\textit{CF}' will rapidly drop under crowded scenes, while the `\textit{8A}' is more stable. Similar issues can be found in \autoref{table:submission} for pose estimation with our proposed w-mAP metric. The above real example illustrates that our proposed metrics can provide a comprehensive evaluation for algorithm, especially for real-world complex events. 

Furthermore, apart from our newly-introduced weighted metrics, we also maintain the original unweighted metrics in our evaluation besides our newly-introduced weighted metrics.  They work together to ensure a comprehensive evaluation in the HiEve dataset.

\begin{table}[t]
	\centering
	\resizebox{\linewidth}{!}{
		\begin{tabular}{c|c|cc|c}
			\hline
			Task                                                                          & Submission name & \multicolumn{2}{c|}{Performance} & Rank \\ \hline
			\multirow{3}{*}{Tracking}                                                     & -               & w-MOTA        & MOTA        & -    \\ \cline{2-5} 
			& `JiaRen.AI'      & 42.93         & 47.40       & 7    \\
			& `Commander'       & 42.47         & 47.41       & 8    \\ \hline
			\multirow{3}{*}{\begin{tabular}[c]{@{}c@{}}Action\\ recognition\end{tabular}} & -               & wf-mAP        & f-mAP       & -    \\ \cline{2-5} 
			& `CF'              & 15.31         & 20.63       & 2    \\
			& `8A'              & 15.09         & 16.25       & 3    \\ \hline
			\multirow{3}{*}{\begin{tabular}[c]{@{}c@{}}Pose\\ Estimation\end{tabular}}    & -               & w-AP          & AP          & -    \\ \cline{2-5} 
			& `Commander'       & 52.25         & 55.47       & 10   \\
			& `DeepBlueAI'      & 52.05         & 56.33       & 11   \\ \hline
	\end{tabular}}
	\caption{Submissions selected from the offical leaderboard on the HiEve website.}
	\label{table:submission}
\end{table}
\section{Conclusion}
We present HiEve, a large-scale dataset for human-centric video analysis. The HiEve dataset covers a wide range of crowded scenes and complex events. We report the results of plenty of approaches in our dataset. Extensive experiments show that the HiEve is a challenging dataset for pose estimation, multi-person tracking, and action recognition. Based on its diverse annotation, we propose two simple baselines, which use cross-annotation information to improve different visual tasks. Experiments on them validate that our HiEve dataset could facilitate multiple visual tasks by diverse annotations.

\section{Declarations}
\textbf{Compliance with Ethical Standards} 
The authors declare no conflicts of interest. All videos in this paper are either collected where the human participants were informed in advance and their consents for data publication were obtained, or obtained from online repositories where the publishing approvals from the video authors were obtained and the human identity information was guaranteed to be properly hidden or blurred.

\noindent \textbf{Data Availability Statement} The datasets analyzed during the current study are all available publicly. Please refer to \href{http://humaninevents.org}{\textcolor{magenta}{http://humaninevents.org}} for further details.

\noindent \textbf{Intended use of HiEve} The authors do not condone with AI systems developed for malicious/unethical surveillance and tracking systems. Any use of the proposed video dataset must adhere to all relevant laws and regulations, including those related to data protection, privacy, and ethical considerations. The proposed video dataset is not to be used for any purpose that violates individual privacy or other legal or ethical standards. The authors are committed to ensuring that the proposed video dataset is used in ways that benefit society and do not cause harm.

\bibliographystyle{spmpsci}      
\bibliography{egbib_hieve}   


\end{document}